\documentclass[sigconf]{acmart}
\usepackage{graphicx}
\usepackage{hyperref}
\usepackage{listings}
\usepackage{booktabs}
\usepackage{multirow}
\usepackage{amsmath}
\usepackage[most]{tcolorbox}
\usepackage{xcolor}
\usepackage{booktabs} 
\usepackage{amsmath}
\usepackage{amsthm}
\usepackage{multirow,tabularx,ragged2e}
\usepackage{adjustbox}

\acmConference[SIGSPATIAL '24]{The 32nd ACM International Conference on Advances in Geographic Information Systems}{October 29-November 1, 2024}{Atlanta, GA, USA}

\newcommand{\GitHubPolData}{\url{{https://github.com/onspatial/pol-outlier-dataset}}}
\newcommand{\GitHubGeoLifeData}{\url{{https://github.com/onspatial/geolife-outlier-dataset}}}
\newcommand{\GitHubProposedApproach}{\url{{https://github.com/onspatial/transferable-outlier-detection}}}

\renewcommand\footnotetextcopyrightpermission[1]{}

\begin{document}

\title{Transferable Unsupervised Outlier Detection Framework for Human Semantic Trajectories}
\author{Zheng Zhang}
\authornote{These authors contributed equally to the paper and are named in random order.}
\orcid{0009-0008-9808-6020}
\affiliation{%
  \institution{Emory University, USA}
  \city{}
  \state{}
  \country{}
}
\email{zheng.zhang@emory.edu}

\author{Hossein Amiri}
\authornotemark[1]
\orcid{0000-0003-0926-7679}
\affiliation{%
  \institution{Emory University, USA}
  \city{}
  \state{}
  \country{}
}
\email{hossein.amiri@emory.edu}

\author{Dazhou Yu}
\orcid{0000-0003-2082-0834}
\affiliation{%
  \institution{Emory University, USA}
  \city{}
  \state{}
  \country{}
}
\email{dazhou.yu@emory.edu}

\author{Yuntong Hu}
\orcid{0000-0003-3802-9039}
\affiliation{%
  \institution{Emory University, USA}
  \city{}
  \state{}
  \country{}
}
\email{yuntong.hu@emory.edu}

\author{Liang Zhao}
\orcid{0000-0002-2648-9989}
\affiliation{%
  \institution{Emory University, USA}
  \city{}
  \state{}
  \country{}
}
\email{liang.zhao@emory.edu}

\author{Andreas Z{\"u}fle}
\orcid{0000-0001-7001-4123}
\affiliation{%
  \institution{Emory University, USA}
  \city{}
  \state{}
  \country{}
}
\email{azufle@emory.edu}

\renewcommand{\shortauthors}{Zhang, Amiri et al.}
\renewcommand{\shortauthors}{Zhang, Amiri et al.}
\begin{abstract}
    Semantic trajectories, which enrich spatial-temporal data with textual information such as trip purposes or location activities, are key for identifying outlier behaviors critical to healthcare, social security, and urban planning. Traditional outlier detection relies on heuristic rules, which requires domain knowledge and limits its ability to identify unseen outliers. Besides, there lacks a comprehensive approach that can jointly consider multi-modal data across spatial, temporal, and textual dimensions. Addressing the need for a domain-agnostic model, we propose the Transferable Outlier Detection for Human Semantic Trajectories (TOD4Traj) framework. TOD4Traj first introduces a modality feature unification module to align diverse data feature representations, enabling the integration of multi-modal information and enhancing transferability across different datasets. A contrastive learning module is further proposed for identifying regular mobility patterns both temporally and across populations, allowing for a joint detection of outliers based on individual consistency and group majority patterns. Our experimental results have shown TOD4Traj's superior performance over existing models, demonstrating its effectiveness and adaptability in detecting human trajectory outliers across various datasets.
\end{abstract}



\maketitle

\section{Introduction}
\label{sec:introduction}

\begin{figure}
    \centering
    \includegraphics[width=1.\linewidth]{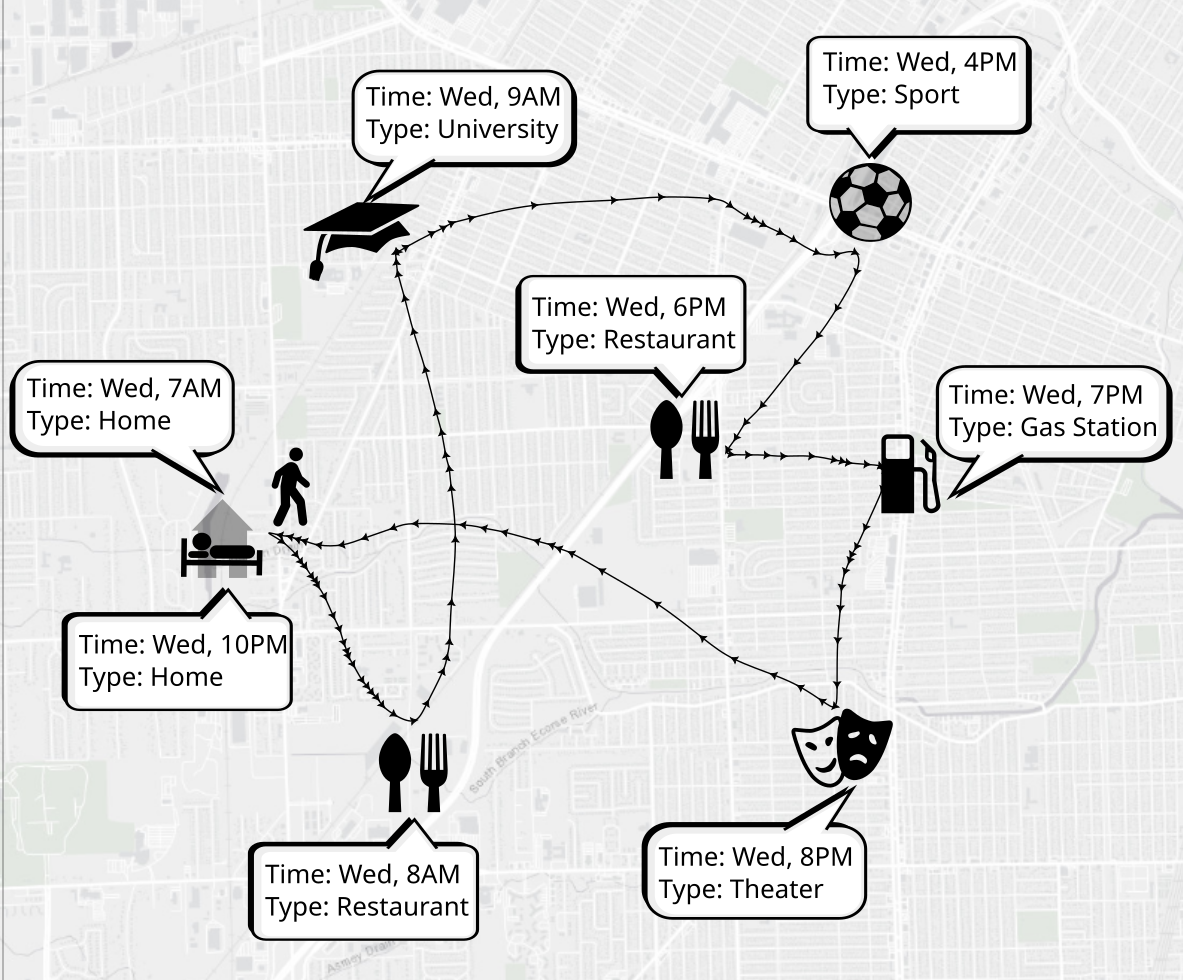}
    \vspace{-4mm}
    \caption{A semantic trajectory of a user including location trajectory and semantic information of points of interest. For each `check-in' location in the figure, there exists an associated text description of location information.}
    \label{fig:semantic_trajectory}
\end{figure}

A semantic trajectory~\cite{parent2013semantic} is a sequence of time-ordered locations where each location is associated with a semantic label like the type of place of interest. A stylized example of a semantic trajectory is shown in Figure~\ref{fig:semantic_trajectory}. It shows a one-day trajectory of a single user starting the day at home and visiting various places of interest (POIs) such as restaurants, a university, and recreational sites. Knowledge discovery in semantic trajectory data has been studied in the past~\cite{alvares2007towards} with a main focus on location prediction~\cite{ying2011semantic}. An important research problem that has received comparatively little attention, due to a lack of available ground truth data, is the problem of outlier detection in semantic trajectories.
%
Yet, detecting semantic trajectory outliers may indicate a change in individual human behavior which has many important applications such as:

\noindent (1) {\textit{Infectious Disease Monitoring}.} A sudden change in behavior such as skipping the sports center or not going to work may indicate that a person is feeling unwell long before severe symptoms arise, infectious disease tests may detect a contagion, and even before the person is consciously aware of feeling unwell themselves. Such information may be leveraged for an early-warning system in cases where the person may have been exposed through a contact-tracing system~\cite{mokbel2020contact,RambhatlaZSSL22,kohn2023epipol}.

\noindent (2) {\textit{Elderly Monitoring}.} GPS-enabled smart-watch technology can be used to monitor the movement of elderly users~\cite{stavropoulos2020iot}. In particular, if the monitored user is showing early signs of dementia, her/his trajectories could show an abrupt change from her/his movement history~\cite{tolea2016trajectory}. Detecting outliers in elder trajectories (and underlying behavior) may thus assist in early-detection and progression-monitoring of dementia.

\begin{figure}[t]
    \centering
    \includegraphics[width=0.98\linewidth, height=2.3in]{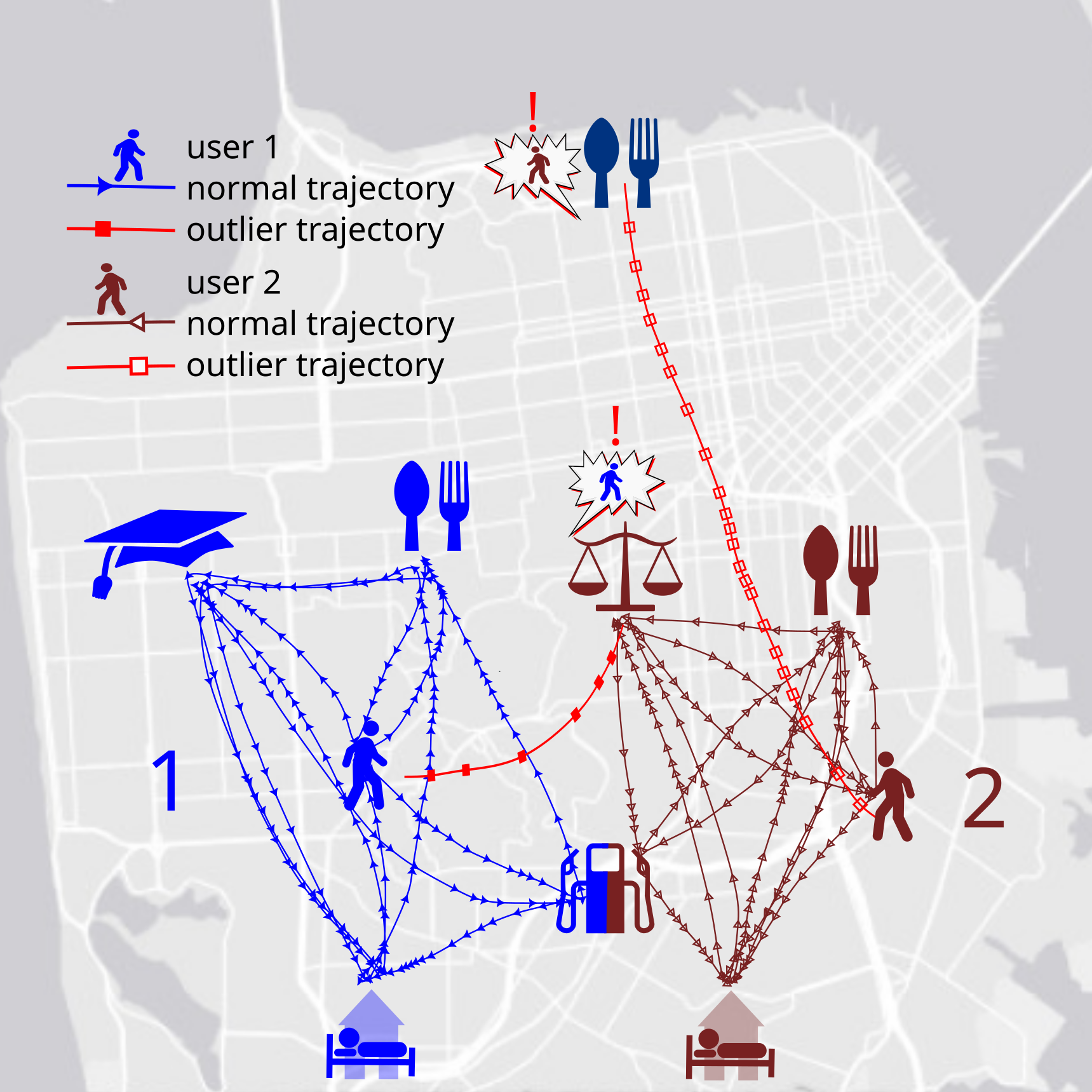}
    \vspace{-0.2cm}
    \caption{An example of spatial and semantic outliers. A spatial outlier for User 2 may be going to a restaurant that is unusually far away. A semantic outlier for User 1 could be going to the same courthouse that User 2 works at. \vspace{-0.3cm} }
    \label{fig:outlier_trajectory}
\end{figure}

What makes semantic trajectory outlier detection a challenging research problem is complexity of humans and their mobility~\cite{mokbel2023towards,mokbel2022mobility} that outliers may have many shapes and forms: Such as spatial outliers of an individual going to unusually distance POIs, temporal outliers of having individuals visit places at unusual times (such as visiting a restaurant in the middle of the night), or semantic outliers (such as an individual who does not normally drink alcohol visit a bar). An additional problem is that ``one person's noise could be another person's signal''~\cite{lee2008trajectory}. To illustrate these challenges, Figure~\ref{fig:outlier_trajectory} shows stylized trajectories of two example users. User 1's normal patterns of life including going to a university in addition to going to his home, nearby restaurants, and a gas station. User 2 lives in a different area and works at a courthouse. An example of a spatial outlier for User 2 may be going to a restaurant that is unusually far away. A semantic outlier for User 1 could be going to the same courthouse that User 2 works at. But since User 1 does not normally go to a courthouse, such a visit could be a deviation from the user's normal patterns of life while the same POI is normal for User 2.

Traditional methods for trajectory outlier detection~\cite{Meng2019,Belhadi2020,basharat2008learning,zhang2012smarter} predominantly rely on heuristic-based rules to identify specific types of outliers, necessitating domain-specific knowledge and limiting the detect of previously unseen outlier behaviors.
Another challenge for semantic trajectory outlier detection is a lack of publicly available datasets. Commonly used (semantic) trajectory datasets such as GeoLife~\cite{zheng2010geolife} trajectories and Location-Based Social Network Check-in Data~\cite{leskovec2016snap} are very sparse, having very few daily trajectories for a specific region or city~\cite{kim2020location} and lack ground truth outlier labels.
Therefore, an open research gap is to transfer an outlier detection model trained on a data-rich city or region (such as a simulated city)
to new regions where no ground truth data is available without compromising performance. Current methodologies frequently employ manually crafted spatial-temporal features, which are usually domain-dependent and lack transfer ability across different domains.

To overcome these limitations, we introduce a \underline{T}ransferable \underline{O}utlier \underline{D}etection framework for Human Semantic \underline{Traj}ectories (TOD4Traj). This framework starts with a modality feature unification module designed to align spatial-temporal and textual data representations. This alignment facilitates the seamless integration of multi-modal data, and enhancing the model's applicability across different datasets.
Additionally, we introduce a unique temporal contrastive learning module designed to represent trajectories by capturing the repetitive nature of mobility patterns. Consequently, outlier degrees are determined by considering both the consistency of an individual's behavior and the prevalent patterns among the majority. To enable other researchers to explore the field of semantic trajectory outlier detection, we make available two types of datasets for benchmarking, including a dataset obtained by systematically including outliers in the GeoLife real-world dataset, and many datasets obtained through a city-level agent-based simulation of patterns of life~\cite{zufle2024silico}. Our experimental findings demonstrate that TOD4Traj substantially surpasses existing models in performance, thereby proving its effectiveness and adaptability in detecting outliers in varied human trajectory datasets.


In general, the contribution of this paper can be summarized into three main points. (1) We proposed a feature-level contrastive learning technique to integrate multi-modal information across spatial, temporal, and semantic dimensions; (2) A trajectory-level contrastive learning module to model the repetitiveness of human mobility patterns; (3) An outlier quantification module to simultaneously measure cross-time and cross-population abnormal behaviors.
The remainder of this work is organized as follows: We begin by discussing existing human semantic trajectory outlier detection algorithms in the Section~\ref{sec:related_works}. This is followed by a formal problem definition and an introduction to the notations in the Section~\ref{sec:preliminary}. Subsequently, in the Section~\ref{sec:methodology}, we delve into the motivation behind our approach and discuss the specific techniques employed. A thorough description of the datasets utilized in our experiments is provided next in the Section~\ref{sec:data}. We conclude with comprehensive experimental results, assessing aspects such as effectiveness, robustness, sensitivity, and efficiency in Section~\ref{sec:results}.

\vspace{-0.35cm}
\section{Related Works}
\label{sec:related_works}
\textbf{Outlier detection in trajectory data.} A crucial aspect of spatio-temporal data analysis, outlier detection is essential for effectively analyzing trajectory information \cite{Gupta2014,liu2024neural}. This technique has seen widespread used in a variety of fields, encompassing applications in wireless sensor networks \cite{Shahid2015, zhang2022unsupervised}, climate monitoring, and transportation management \cite{Meng2019, Wang2020}.
Surveys of traditional trajectory outlier detection algorithms can be found in~\cite{Meng2019,Belhadi2020}. 
Important examples of such algorithms include \cite{Su2023} where the authors us a transfer learning approach to find outliers in areas where only a small set of trajectories are observed.
In \cite{Daneshpazhouh2014}, the authors propose an entropy-based method designed specifically for outlier detection in scenarios where the training data contains only a few positive instances. 
In \cite{Shi2023}, a real-time urban traffic outlier detection system that leverages both individual and group outlier detection was proposed.
However, these approaches all aim at finding outliers in traditional trajectories defined by sequences of geo-locations without using any semantic information of the visited locations.

\textbf{Contrastive learning} has emerged as a promising technique in the field of unsupervised representation learning \cite{hadsell2006dimensionality}. The core idea behind contrastive learning is to exploit the relationships between samples to learn meaningful representations. By contrasting positive pairs (similar samples) with negative pairs (dissimilar samples), it aims to map similar samples closer in the latent space while pushing dissimilar samples further apart. This approach obviate the need for explicit annotations or labels, making it particularly suitable for scenarios with limited labeled data. Numerous contrastive learning methods have been proposed, such as InfoNCE \cite{oord2018representation}, SimCLR \cite{chen2020simple}, and MoCo \cite{he2020momentum}. These methods have demonstrated impressive results in various domains, including computer vision and natural language processing, showcasing that contrastive learning as a powerful tool for unsupervised representation learning. However, the exploration of contrastive learning in the domain of semantic trajectories remains largely unexplored due to the inherent complexity and unstructured nature of trajectory data.

\textbf{Semantic Trajectory Representation} methods can be grouped into 1) location-level semantic information~\cite{chen2020parallel,cong2012efficient,zheng2015approximate,zheng2017popularity} which associate each visited point of interest (or staypoint) with semantic information and 2) trajectory-level semantic information~\cite{shang2012user,liu2013moir} which associate an entire trajectory with a semantic label. Our approach uses the more general cases of location-level semantic information. Existing work on semantic trajectories has tackled important tasks such as semantic trajectory prediction~\cite{ying2011semantic,yao2017serm} and clustering~\cite{liu2020stccd}. However, to the best of our knowledge, no work has tackled the problem of finding outliers in semantic trajectories. One possible reason for the lack of existing research in this field is the lack of semantic trajectory data that includes outlier information. In this work, we fill this gap by 1) creating simulated semantic trajectory datasets where outlier information is directly included in the semantic trajectory generation, 2) providing a real-data set of semantic trajectory outliers based on the existing GeoLife~\cite{zheng2010geolife} data, and 3) proposing a first approach towards outlier detection in semantic trajectories.

\vspace{-0.4cm}
\section{Preliminaries}
\label{sec:preliminary}

A semantic trajectory of an individual user can be represented as a sequential list of staypoints denoted by $\mathcal{T}=\{\mathbf{p_1}\rightarrow \mathbf{p_2}\rightarrow \dots\rightarrow \mathbf{p_n}\}$, where each staypoint $\mathbf{p_i}=(s_i; t_i; c_i)$ includes a spatial coordinate $s_i=(x_i, y_i)$, a timestamp $t_i$, and a semantic location class $c_i$. Here, $n$ is the total count of staypoints in a trajectory. The spatial coordinates $s_i$ specify the longitude $x_i$ and latitude $y_i$ positions, while the semantic class $c_i$ identifies the type of location, such as restaurant or apartment, through descriptive text. A sub-trajectory of $\mathcal{T}$, denoted as ${T}^{(i,j)}\subseteq \mathcal{T}$, can be formally defined as a contiguous segment of staypoints from $\mathcal{T}$. This subset is represented as ${T}^{(i,j)} = \{\mathbf{p_i}\rightarrow \mathbf{p_{i+1}}\rightarrow \dots\rightarrow \mathbf{p_j}\}$, where $1\leq i \leq j \leq n$ and $i,j$ are indices within the original sequence $\mathcal{T}$. This definition captures a portion of the user’s trajectory, maintaining the chronological and spatial integrity of the original sequence.
To encompass the collection of trajectories from multiple users, let $\mathcal{U}$ be the set of all users. We denote the entire set of all users as a database $\mathcal{DB}=\{\mathcal{T}_1,\mathcal{T}_2,\dots,\mathcal{T}_{|\mathcal{U}|}\}$, where $|\mathcal{U}|$ denotes the total number of users. Thus, each $\mathcal{T}_u\in\mathcal{DB}$ represents a sequence of semantic trajectories of a distinct user $u$.


%
Given the above definitions, here we formally formulate the semantic trajectory outlier detection problems:

{\noindent \textbf{Problem 1. Cross-Time Semantic Trajectory Outlier Detection.}}
\textit{Given a user \( u \) from the user set \( \mathcal{U} \) and their set of trajectories \( \mathcal{T}_u \) in database \( \mathcal{DB} \), the task is to identify outlier trajectories \( T_{outlier} \subseteq \mathcal{T}_u \) that exhibit significant deviation from the user's typical trajectory patterns over different time periods. These deviations are quantified using a score function \( f_t \), which measures the degree of outlierness relative to the user's historical trajectory data.}

{\noindent \textbf{Problem 2. Cross-Population Semantic Trajectory Outlier Detection.}}
\textit{For each user \( u \in \mathcal{U} \), let \( \mathcal{T}_u \) represent their set of trajectories. The task involves identifying outlier trajectories \( T_{outlier} \subseteq \mathcal{T}_u \) that diverge significantly from the majority pattern set \( \mathcal{M} \), derived from aggregating trajectories across all users in \( \mathcal{U} \). Outlier detection is based on a score function \( f_p \), which evaluates the extent of deviation from common patterns observed across the population.}

This goal presents several unique challenges:
(1) \textbf{Difficulty in seamlessly integrating multi-modal information across spatial, temporal, and semantic dimensions.}
Each modality carries unique and critical information about user behaviors and patterns. Considering the interactions among data from various modalities is essential for a comprehensive identification of complex outliers.
(2) \textbf{Difficulty in tracking temporal shift in user behaviors.} Human behavior is dynamic and can change due to numerous factors such as personal preferences, environmental changes, and social influences. Existing methods typically use rule-based methods, which is insufficient to handle unseen pattern shifts. Capturing these evolving patterns over time, especially in a way that accurately reflects significant shifts, demands advanced modeling techniques.
(3) \textbf{Difficulty in analyzing varied user behaviors across populations.} An outlier may also occur when the trajectory pattern of an individual diverges significantly from the majority pattern observed across the broader population. The difficulty in detecting such outliers stems from the variability in behavior patterns and scalability issues with algorithms.

\vspace{-0.4cm}
\section{Methodology}
\label{sec:methodology}

\begin{figure*}[t]
    \centering
    \includegraphics[width=.79\linewidth]{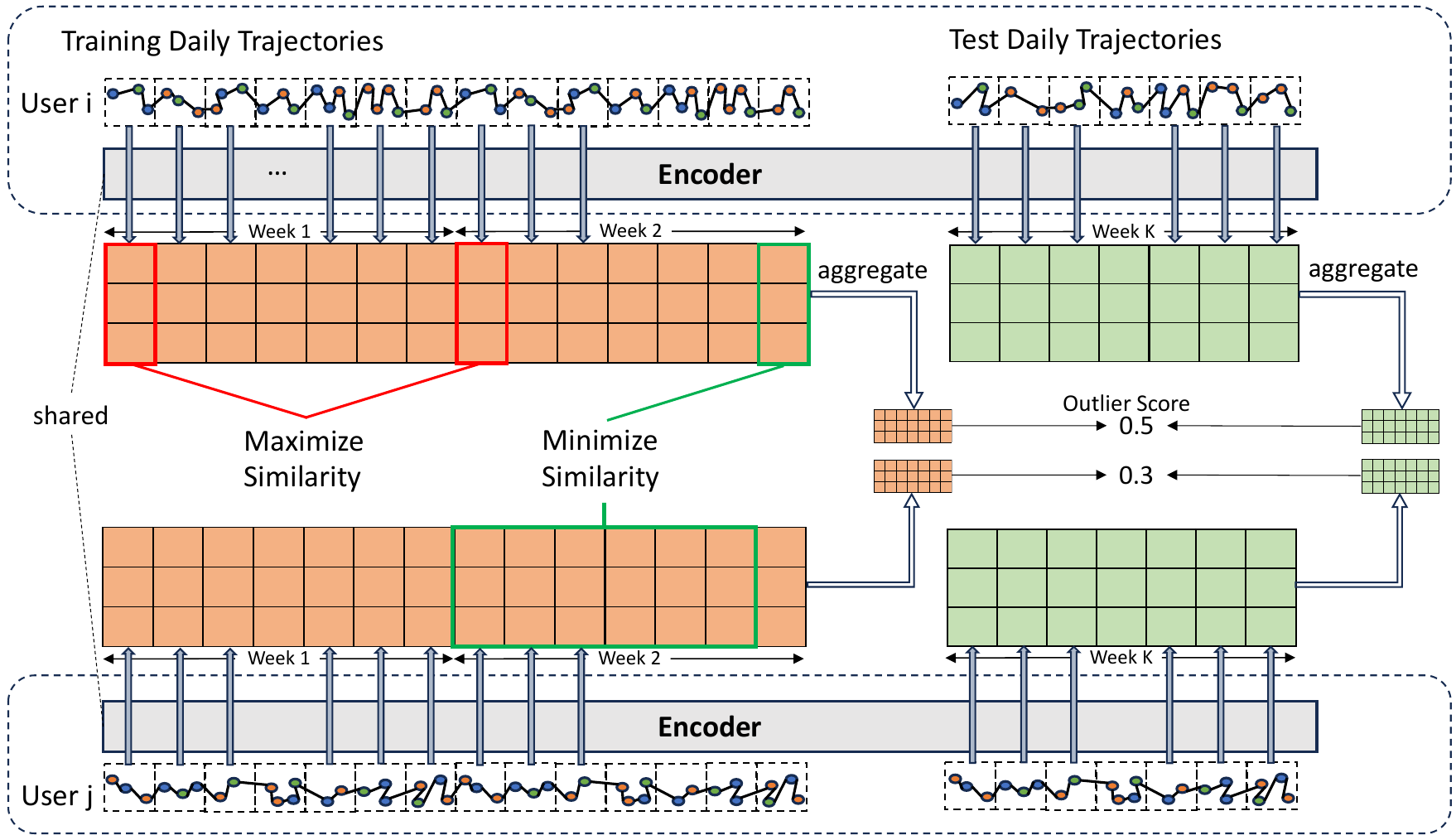}
    \vspace{-0.3cm}
    \captionsetup{width=1.\linewidth}
    \caption{The illustration of our proposed model framework. (Top \& Bottom) Extraction of spatio-temporal semantic embeddings from input trajectories;  (Middle left) Contrastive learning based on human semantic trajectory periodicity; (Middle right) Quantification of trajectory anomaly scores based on dissimilarity between train and test embeddings.}
    \label{fig:framework}
    \vspace{-0.1cm}
\end{figure*}

In this section, we propose Transferable Outlier Detection for Human Semantic Trajectories (TOD4Traj) framework. Notably, our method can identify outlier behaviors without the need for labeled data. Our framework is composed of three modules:
(1) To integrate the modality features across spatial, temporal and semantic dimensions, we developed a Spatial Temporal-to-Semantic contrastive learning strategy that aligns representations from disparate sources into a unified feature space, enhancing the detection of joint anomalies. Furthermore, by aligning spatial-temporal information with semantic data, we facilitate the transferability of spatial-temporal features across different datasets;
(2) To effectively monitor changes in user behaviors over time, we employ a temporal contrastive learning approach that identifies the repetitiveness of human mobility patterns. This technique, produces trajectory-level embeddings, seamlessly merging spatial-temporal and semantic data from trajectory sequences. 
(3) To identify abnormal trajectory patterns across users' behaviors, we have implemented a comparative analysis framework that leverages population-wide mobility trends. This framework compares individual trajectories with collective population behaviors, allowing for the detection of outliers that diverge from majority patterns.


\subsection{Modality Alignment through Spatial Temporal-to-Semantic Contrastive Learning}\label{section:modality}
In order to fully exploit the compatibility between semantic information and spatio-temporal information carried by the semantic information, it is crucial to enable cross-modal alignment. In this section, we developed a Spatial-Temporal to Semantic Contrastive Learning module, which aimed at integrating various data modalities. The core concept involves identifying the co-occurrence patterns within different modalities as observed in semantic spatial-temporal trajectories. Through this approach, we achieve the alignment of data across modalities into a unified, semantically-enriched, high-dimensional embedding feature space.

To effectively align spatio-temporal information with semantic information into a unified, semantically-enriched, high-dimensional embedding feature space, we utlize the co-occurrence patterns within different modalities as observed in trajectories. We adopt natural language of semantic information as supervision labels, leveraging its distinct advantages over other data sources. The primary aim of this technique is to learn a mapping that converts spatio-temporal information into natural language semantics embeddings, thereby harnessing their inherent coexistence pattern. For example, this technique aims to closely associate time-specific phrases like ``Friday 6PM'' with contextually relevant semantic labels, such as ``entertainment place''. Similarly, it seeks to connect physical locations visited with their semantic significance, enhancing the model's ability to interpret and utilize the feature embeddings meaningfully.
Specifically, the Spatial-Temporal to Semantic Contrastive Learning module is designed to align the spatial-temporal representation with the semantic representation of the same trajectory point, aiming to maximize their mutual information. This process involves enhancing the similarity between spatial-temporal and semantic representations of a positive pair relative to that of negative pairs.

Formally, given a user \(u\) from the user set \(\mathcal{U}\) and its set of trajectories \(\mathcal{T}_u\), we explicitly link each spatial-temporal information $s_i, t_i$ with its corresponding semantic class $c_i$ within the same staypoint $\mathbf{p_i}$. Thus, we define the positive set $\mathcal{P}$ as:
\begin{equation}
    \mathcal{P}(u) = \{ (s_i, t_i; c_i) \mid s_i, t_i, c_i \in \mathbf{p_i}, \forall \mathbf{p_i} \in \mathcal{T}_u \},
\end{equation}

and the negative set contains the pairing of a spatial-temporal information $s_i, t_i$ with the semantic class $c_j$ from a different staypoint, which can be defined as:

\begin{equation}
    \mathcal{N}(u) = \{ (s_i, t_i; c_j) \mid s_i, t_i \in \mathbf{p_i}, c_j \in \mathbf{p_j}, i \neq j, \mathbf{p_i}, \mathbf{p_j} \in \mathcal{T}_u \}.
\end{equation}

To encourage the similarity between the positive pairs and dissimilarity between negative pairs, we introduce the following contrastive learning objective function:

\begin{align}
    \begin{split}
        \mathcal{L}_{\mathrm{Align}} = -  \sum_{(s_i, t_i; c_i) \in \mathcal{P}(u)} & \log \frac{e^{\text{sim}(\mathbf{d_{c_i}}, \mathbf{d_{t_i}}) / \tau}}{\sum_{(s_j, t_j; c_j) \in \mathcal{P}(u) \cup \mathcal{N}(u)} e^{\text{sim}(\mathbf{d_{c_i}}, \mathbf{d_{t_j}}) / \tau}} \\
        + \quad & \log \frac{e^{\text{sim}(\mathbf{d_{c_i}}, \mathbf{d_{s_i}}) / \tau}}{\sum_{(s_j, t_j; c_j) \in \mathcal{P}(u) \cup \mathcal{N}(u)} e^{\text{sim}(\mathbf{d_{c_i}}, \mathbf{d_{s_j}}) / \tau}},
    \end{split}
\end{align}
where $\mathbf{d_{c_i}}$ is the embedding from the text encoder for the semantic class $c_i$, and $\mathbf{d_{s_i}}$ and $\mathbf{d_{t_i}}$ are the embeddings produced by the spatial-temporal encoder for the spatial-temporal information $s_i$ and $t_i$, respectively. The function $\text{sim}(\cdot, \cdot)$ computes the similarity between pair embeddings from different modalities, and $\tau$ is a temperature scaling parameter that controls the separation of distributions.

In more details, a frozen pre-trained language model (e.g. BERT) is utilized to project textual information $c_{ij}$ into a vectorized text representation: $\mathbf{d}_{c_{i}} = \mathrm{PLM}(c_{i})$,
where $\mathbf{d}_{c_{ij}}$ is the hidden representation of the $\mathrm{[CLS]}$ token computed from the last layer of the $\mathrm{PLM}$ (pretrained language model) encoder.

To align the spatial temporal information to the semantic embeddings from $\mathrm{PLM}$ encoder, we develop two learnable mapping modules $\mathcal{M}_{s}$ and $\mathcal{M}_{t}$ to transform the spatial and temporal information to the semantic embedding space. Specifically, the transformed temporal embeddings can be represented as $\mathbf{d}_{t_{i}} = \mathcal{M}_{t}(t_{i})$, and the transformed spatial embeddings can be represented as $\mathbf{d}_{s_{i}} = \mathcal{M}_{s}(s_{i})$.

\textbf{Why adopting natural language embeddings as the supervision labels?}
An advantage of adopting natural language supervision is because it is inherently scalable and easily interpretable by humans, unlike spatiotemporal labels. This approach also benefits from the rapid advancement in language models, enabling the generalization to unseen labels. For example, through our approach, once the model learns to associate a specific time stamp, such as 6 pm, or a defined area with certain locations like McDonald's, it gains the ability to generalize this knowledge. Consequently, the model can recognize and associate similar time or region patterns with other fast-food stores like Burger King without requiring further specific training. This method not only simplifies the learning process but also enhances the model's ability to apply learned concepts to new, yet related scenarios, thereby increasing its effectiveness in understanding and interpreting complex spatiotemporal and semantic relationships.

\subsection{Modeling Semantic Trajectories through Regular Pattern Contrastive Learning} \label{section:regular}

After aligning the embedding vectors to encapsulate both spatial-temporal and semantic data, we focus on developing a framework that aggregates these point-level embeddings into coherent trajectory-level representations. A key observation in human mobility patterns is the temporal consistency in an individual's activities~\cite{gonzalez2008understanding}. This consistency is evident in the recurring nature of activities and mobility patterns on specific days of the week, mirroring similar behaviors on equivalent days in history. For instance, the behavior of a user on the current workday is often similar to the activities performed on previous workdays. This self-consistency extends to other days of the week as well.

Recognizing this consistent nature, we are inspired to incorporate two self-supervised learning tasks into our framework: (a) classifying if two daily trajectories were generated by the same user, and (b) classifying if two trajectories of the same user were generated on the same days of the week. The intricacies of this approach are depicted in Figure~\ref{fig:framework}, which illustrates how the consistency in trajectories is leveraged to enhance the learning process.

Formally, large trajectories of each user are segmented into individual daily trajectories. For each user $u$ we observe a set of trajectories $\mathcal{T}_u=[T_{u,1},...,T_{u,D_u}]$ where each trajectory $T_{u,d}\in\mathcal{T}_u$ corresponds to the $d$'th daily trajectory of the user $u$ and $D_u$ denotes the number of daily trajectories observed for the user $i$.  Each daily trajectory, denoted by \(T_{u,d}, d \in [1,D_u]\), is constituted by the unique locations visited by the user $u$ within the date \(D_u\).

To generate daily-level embeddings from distinct spatial, temporal, and semantic embeddings, we employ a deep sequential encoder \(f: T \rightarrow \mathbf{z} \) to map each daily trajectory $T_{u,d}\in\mathcal{T}_u\in\mathcal{DB}$ of the user $u$ into latent high-dimensional embeddings. This stage involves the extraction of sequentially organized spatio-temporal-semantic information from each daily trajectory, transforming them into meaningful representations. We present our model as a general framework, accommodating various commonly employed deep sequential models as potential encoders. This flexibility allows for the incorporation of classical models such as Recurrent Neural Networks (RNN) or modern architectures like Transformers. In our experimental evaluation, we explore different encoder models to verify the framework generalizability.

Finally, as illustrated in the left (green shade) of Figure~\ref{fig:framework}, consider a set of daily trajectories $\mathcal{T}_u$ belonging to the user \(u\). For a specific daily trajectory $T_{u,d}\in \mathcal{T}_u$, we construct the positive set of samples with other days corresponding to the same intrinsic pattern (e.g. a working Monday). For notational simplicity, we denote the set of days sharing the same pattern as
\begin{equation}
    \mathcal{D}(d) = \{d + fq | q \in \mathbb{Z} \setminus \{0\}, 1 \leq d + fq \leq D_u\},
\end{equation}
where $f$ is the frequency of repeating the same pattern (e.g. $f=7$ for a weekly repetition). Therefore, the positive pairs set $\mathcal{S}$ can be denoted as
\begin{equation}
    \mathcal{S} (T_{u,d}) = \{T_{u,d^{\prime}} | d^{\prime}\in \mathcal{D}(d)\}
\end{equation}

Conversely, for constructing negative pairs, we sample from other users and weekdays that do not align with the target day, which is denoted as
\begin{equation}
    \mathcal{I} (T_{u,d}) = \{T_{v,d^{*}} | v \in\mathcal{V}, v\neq u , d^{*}\notin \mathcal{D}(d)\}
\end{equation}

To operationalize the contrastive learning, without loss of generality, the objective function can be written as:
\begin{align}
    \mathcal{L}_{\mathrm{Consistency}} & = - \sum_{u\in \mathcal{U}} \sum_{d\in [1,D_u]} \log \frac{ s_{\text{pos}}(T_{u,d}) }{s_{\text{pos}}(T_{u,d}) + s_{\text{neg}}(T_{u,d})},\notag
\end{align}
\begin{align}
    \begin{split}
        s_{\text{pos}}(T_{u,d}) & = \sum_{T_{u,d^{\prime}}\in \mathcal{S}(T_{u,d})} e^{ \text{sim}(\mathbf{z}({T_{u,d}}), \mathbf{z}({T_{u,d^{\prime}}) })/\varsigma} \\
        s_{\text{neg}}(T_{u,d}) & = \sum_{T_{u,d^{*}}\in \mathcal{I}(T_{u,d})} e^{ \text{sim}(\mathbf{z}({T_{u,d}}), \mathbf{z}({T_{u,d^{*}}) })/\varsigma}
    \end{split}
\end{align}
where $\mathbf{z}(\cdot)$ denotes the daily-level trajectory embeddings, and $\varsigma$ is the temperature parameter. The negative samples are chosen from the embeddings corresponding to different users and different days of the week, thereby ensuring the maximization of dissimilarity among the selected negative samples.

We acknowledge that this self-supervised approach which uses the task of classifying whether two trajectories 1) belong to the same user and 2) from the same day-of-the-week may incur confusion in for special cases such as holidays, where a holiday-Thursday may be more similar to a Sunday for some users. However, the dissimilarity between users should remain high, as different users will have different home locations, different work locations (during work days), and different favorite locations. Thus, there should still be substantial contrast between a positive sample that suffers from confusion due to holidays and a negative sample between different users (and different days-of-the-week).

\subsection{Quantification of Outlier Scores} \label{quatification}
Given the trained contrastive model that can extract human mobility pattern behavior from both spatial-temporal and semantic information, we now focus on the process of quantifying the degree of being abnormal. It is important to recognize that outliers may occur both cross-time or cross-population, as defined in Section~\ref{sec:preliminary}. Intuitively, a cross-time outlier is indicated when a user's current trajectory pattern significantly deviates from their historical patterns. Conversely, a cross-population outlier is suggested when this pattern markedly differs from those of whole populations. This dual-focus analysis allows for a comprehensive understanding of deviations in mobility behavior, while existing methods~\cite{basharat2008learning, zhang2012smarter, liu2020online, han2022deeptea} typically only focus on one kind of the outliers.

To detect both cross-time and cross-population outliers, a straightforward approach involves comparing the user's current trajectory embedding with their past trajectory embeddings and those of other users. However, this method faces significant challenges. Measuring global mismatches comprehensively would necessitate calculating the pairwise similarity for every user pair, leading to a quadratic increase in computational complexity. To circumvent this issue, we suggest leveraging the \textit{low-rank} properties of human mobility patterns for measuring outliers. Typically, human mobility patterns exhibit low-rank characteristics in large user sets, attributed to the regularity of human behaviors. Individuals generally adhere to a limited range of routines and visit a restricted set of locations consistently, resulting in repetitive movement patterns across a broad population. This uniformity means the entire dataset of human movements can be effectively summarized by a small set of core factors or dimensions, reflecting its low-rank nature. 

Unfortunately, applying traditional low-rank techniques like Singular Value Decomposition (SVD) directly to this problem introduces generalization issues with new data, making it unsuitable for online detection methods. Furthermore, SVD demands considerable computational resources, presenting a significant challenge for efficient implementation.

To effectively harness the low-rank property within the entire dataset of human movement trajectories, we introduce a soft clustering objective into our overall training objective function. By optimizing a small set of clustering centroids, we aim to capture the essence of low-rank movement patterns. Consequently, the proximity of each trajectory to its nearest clustering centroid serves as a measure of its deviation from the mainstream patterns in the dataset. This distance becomes a crucial indicator for assessing the degree of abnormality, with greater distances suggesting more significant deviations from typical movement behaviors.

Formally, given a set of $K \ll |\mathcal{U}|$ learnable centroids $\{\mathbf{b}_k|k\in[1,K]\}$, the soft clustering objective function can be written as:
\begin{align}
    \begin{split}
        \mathcal{L}_{Clustering} &= \sum_{u\in\mathcal{U}} \sum_{d\in[1,D_u]} \sum_{k\in[1,K]} \delta_{u,d}^{(k)} \ell(\mathbf{z}(T_{u,d}), \mathbf{b}_k),\\
        \delta_{u,d}^{(k)} &= \frac{\ell(\mathbf{z}(T_{u,d}), \mathbf{b}_k)}{\sum_{k\in[1,K]}\ell(\mathbf{z}(T_{u,d}), \mathbf{b}_k)}
    \end{split}
\end{align}
where $\ell$ represents a distance measurement function, typically selected as $\|\cdot\|^2$. Here $\delta_{u,d}^{(k)}$ signifies the coefficient weight that allocates the current embedding $\mathbf{z}(T_{u,d})$ to the $k$-th centroid $\mathbf{b}_k$.

Therefore, the overall training objective can be written as:
\begin{equation}
    \mathcal{L} = \mathcal{L}_{\mathrm{Consistency}} + \beta \mathcal{L}_{\mathrm{Clustering}},
\end{equation}
where $\beta$ is a hyperparameter to balance between two terms.

Finally, the quantification of outlier scores for both cross-time and cross-population anomalies is achieved by assessing the discrepancies between (1) the historical and current patterns of an individual user, and (2) the current pattern of a user and all centroids.
To achieve this, we first divide the historical trajectory data of a user into sets corresponding to each day pattern, using the notation \(\mathcal{D}(d)\) as defined earlier. For each day \(d\), we compute the average embedding of the historical trajectories as $$\mathbf{h}_{u,d} = \frac{1}{|\mathcal{D}(d)|} \sum_{d' \in \mathcal{D}(d)} \mathbf{z}(T_{u,d'}).$$
Similarly, we compute the average embeddings $\hat{\mathbf{h}}_{u,d}$ for each day pattern from the new incoming trajectory data in the same way. Then, the cross-time outlier score for the current trajectory data can be quantified by measuring the dissimilarity between the historical embedding \(\mathbf{h}_{u,d}\) and the current embedding \(\hat{\mathbf{h}}_{u,d}\):
\begin{equation}
    \text{Cross-Time}({u}) = 1 - \frac{1}{f}{\sum_{d\in[1,f]}\mathrm{sim}(\mathbf{h}_{u,d}, \hat{\mathbf{h}}_{u,d})},
\end{equation}
where  $f$ is the total number of days in the considered period (e.g., $f=7$ for a week). Similarly, the cross-population outlier score can be quantified by measuring the dissimilarity between the current embedding with the closest centroid:
\begin{equation}
    \text{Cross-Population}({u}) = \max\{1 - \mathrm{sim}(\hat{\mathbf{h}}_{u,d}, \mathbf{b}_k) | k\in[1,K]\}.
\end{equation}

\section{Experimental Results}
\label{sec:results}

We implemented all the methods, including our proposed method and competitor methods, through the Pytorch Framework. We have open-sourced all the code in the supplementary material. For a fair comparison, we require all models to follow the same experimental settings and data splits.
All methods, including our proposed method and those of competitors, were implemented using the PyTorch Framework. In an effort to support transparency and reproducibility in the research community, we have provided all corresponding code at \GitHubProposedApproach.
For a fair comparison, we maintained consistent experimental conditions across all models.

\vspace{-0.3cm}
\subsection{Experimental Datasets}
\label{sec:data}
The datasets used for this research include six simulated
datasets using the Agent-Based Patterns-of-Life Simulation~\cite{zufle2023urban,kim2020location,amiri2024patterns} and one real-world dataset based on the GeoLife dataset~\cite{zheng2010geolife}.
Specifications of the datasets, including details and key attributes, can be found in Table \ref{table:dataset}. 
The source code of the simulation and data processing of the GeoLife dataset is accessible through the GitHub repositories:  \GitHubPolData ~and~ \GitHubGeoLifeData, respectively. In addition, all datasets are available for download at https://osf.io/rxnz7/ and described in \cite{amiri2023massive,zhang2023large}.

\subsubsection{Agent-Based Simulation of Patterns of Life }\label{app:pol}

The patterns of life simulation was designed to emulate human needs and behavior in an urban environment~\cite{zufle2023urban}. 
Within the simulated environment, virtual entities referred to as agents perform actions that mirror human activities. These include attending work, forming friendships, engaging in social gatherings, and more. The agents' existence is crafted to resemble human life in a real-world environment (roads, buildings) obtained from OpenStreetMap~\cite{bennett2010openstreetmap,atwal2022predicting}. Throughout their simulated lives, agents navigate to diverse locations, including restaurants, workplaces, residential apartments, and recreational venues.
A salient feature of the simulation is the generation of comprehensive log files. These logs contain extensive data regarding the agents, including their location and current state information, thus allowing for in-depth analysis and research.

In our study, we generated data by running simulations over four distinct maps, namely Fairfax County, Virginia, USA (FVA); the French Quarter of New Orleans, Louisiana, USA (NOLA); Atlanta, Georgia, USA (ATL); and Beijing, China (BJNG). The simulations were conducted over a period of 450 days to replicate normal life, followed by an additional 14 days to incorporate abnormal behavior into the regular patterns.
We introduced three specific types of abnormal behavior that define outliers trajectories:

\begin{itemize}
    \item \textbf{Hunger outlier:} An agent under this category becomes hungry more quickly. Such agents have to go to restaurants or their homes much more often.
    \item \textbf{Social outlier:} This type of agent randomly selects recreational sites to visit when needed, rather than being guided by their attributes and social network.
    \item \textbf{Work outlier:} Agents in this category abstain from going to work on workdays.
\end{itemize}

We further divided these abnormalities into three intensity levels: red, orange, and yellow. Red outliers exhibit extremely abnormal behavior, orange outliers act moderately abnormal, and yellow outliers display abnormal behavior less frequently. For example, a work outlier will decide not to go to work 100\%, 50\%, or 20\% of the time when classified as red, orange, or yellow, respectively. We divide the simulation into 450 simulation of days of normal behavior followed by 14 days of a small number of agents exhibiting outlier behavior. Details can be found in Table~\ref{table:dataset} and, an extended version of the dataset can be found in \cite{amiri2024urban}.

\subsubsection{Real World Dataset}\label{app:geolife}

The real-world dataset for this study was created using the Microsoft Research Asia's GPS Trajectory dataset~\cite{zheng2010geolife}. Since the original data did not conform to a check-in format, we employed the method outlined in \cite{zheng2008mining} to extract stay points, thereby transforming the data to fit the check-in pattern used in life simulation studies.
Next, we utilized OpenStreetMap to categorize locations into four groups: apartments, workplaces, pubs, and restaurants. Given that OpenStreetMap encompasses a broad array of categories and types, we manually classified them into these four distinct groups.
Upon preprocessing the data, we eliminated agents with fewer than 50 records, resulting in a final count of 69 agents with a total of 14,080 training trajectories and 3,552 test trajectories. Within the context of the GeoLife dataset, we introduced a specific outlier type called the ``imposter outlier". An agent acting as an imposter outlier by switching the trajectories with another agent after a specific time point. 
The dataset was then divided into two segments: 80\% of the stay points for training and introduced outliers into the remaining 20\% for test. 

\begin{table}[t]
    \centering
    \begin{adjustbox}{width=1.0\columnwidth,center}
        \begin{tabular}{|c|c|c|c|c|}
            \hline
            Outlier Type & \#Agents & Source  & Period      & \#Outliers \\
            \hline
            hunger       & 1000     & POL     & 450+14 days & 90         \\
            \hline
            work         & 1000     & POL     & 450+14 days & 30         \\
            \hline
            social       & 1000     & POL     & 450+14 days & 30         \\
            \hline
            combined     & 3000     & POL     & 450+14 days & 150        \\
            \hline
            imposter     & 69       & GeoLife & 4 years     & 20         \\
            \hline
        \end{tabular}
    \end{adjustbox}
    \caption{Detailed statistical information of the datasets utilized in this paper. Here `POL' donotes Pattern-of-Life data.\vspace{-0.5cm}}
    \label{table:dataset}
\end{table}

\subsection{Experimental Settings}

\begin{table*}[t]
    \centering
    \begin{adjustbox}{width=.96\textwidth,center}
\begin{tabular}{@{}lcccccccc@{}}
\toprule
 & \multicolumn{4}{c}{ATL} & \multicolumn{4}{c}{NOLA} \\
\cmidrule(lr){2-5} \cmidrule(l){6-9}
Model & Top-10 Hits & Top-100 Hits\textsuperscript{*} & AP score & AUC score & Top-10 Hits & Top-100 Hits & AP score & AUC score \\
\midrule
OMPAD & 0 & 7 & 0.0571 & 0.5257 & 2 & 10 & 0.0776 & 0.5968 \\
MoNav-TT & 1 & 5 & 0.0893 & 0.4863 & 1 & 3 & 0.0503 & 0.5026 \\
TRAOD & 1 & 10 & 0.0582& 0.5018 & 0 & 4 & 0.0485 & 0.5011 \\
DSVDD & 5 & 36 & 0.2601 & 0.5835 & 9 & 28 & 0.2093 & 0.5829 \\
DAE & 3 & 17 & 0.0962 & 0.5465 & 1 & 7 & 0.0648 & 0.5885 \\
GM-VSAE  & 4   & 29      & 0.1987      & 0.5564          & 5           & 20           & 0.1786          & 0.5672                \\
DeepTEA  & 5    & 26     & 0.2008                   & 0.6012          & 5           & 26           & 0.2186          & 0.6395     \\
\midrule
Ours-MLP & \textbf{10} & 34 & 0.2782 & 0.6824 & \textbf{10} & 41 & 0.3376 & 0.6985 \\
Ours-RNN & \textbf{10} & 32 & 0.2780 & 0.6233 & \textbf{10} & 27 & 0.2325 & 0.5940 \\
Ours-CNN & \textbf{10} & \textbf{42} & \textbf{0.3205} & \textbf{0.7215} & \textbf{10} & \textbf{46} & \textbf{0.3631} & \textbf{0.7185} \\
Ours-Transformer & \textbf{10} & 34 & 0.2436 & 0.6735 & \textbf{10} & 34 & 0.2903 & 0.6970 \\
\midrule
 & \multicolumn{4}{c}{FVA} & \multicolumn{4}{c}{BJNG} \\
\cmidrule(lr){2-5} \cmidrule(l){6-9}
OMPAD & 0 & 4 & 0.0598 & 0.5322 & 1 & 9 & 0.0704 & 0.5655 \\
MoNav-TT & 0 & 0 & 0.0501 & 0.5014 & 1 & 5 & 0.0893 & 0.4863 \\
TRAOD & 0 & 7 & 0.0515& 0.5090 & 0 & 6 & 0.0553 & 0.5169 \\
DSVDD & 5 & 26 & 0.2166 & 0.5995 & 10 & 29 & 0.2155 & 0.5643 \\
DAE & 1 & 7 & 0.0569 & 0.5138 & 0 & 10 & 0.0671 & 0.5568 \\
GM-VSAE  & 4           & 22           & 0.1534          & 0.5859       & 4           & 16           & 0.1068          & 0.5479          \\ 
DeepTEA  & 5           & 30           & 0.2221          & 0.6182      & 6           & 24           & 0.2084          & 0.5873          \\
\midrule
Ours-MLP & \textbf{10} & 32 & 0.2509 & 0.6561 & \textbf{10} & 34 & 0.2800 & 0.6587 \\
Ours-RNN & \textbf{10} & 27 & 0.2325 & 0.5940 & \textbf{10} & 31 & 0.2573 & 0.6065 \\
Ours-CNN & \textbf{10} & \textbf{40} & \textbf{0.3151} & \textbf{0.6669} & \textbf{10} & \textbf{66} & \textbf{0.4899} & \textbf{0.7513} \\
Ours-Transformer & \textbf{10} & 33 & 0.2171 & 0.6628 & \textbf{10} & 33 & 0.2499 & 0.6219 \\
\midrule
 & \multicolumn{4}{c}{Geolife} & \multicolumn{4}{c}{ATL-Large} \\
\cmidrule(lr){2-5} \cmidrule(l){6-9}
OMPAD & 1 & 4 & 0.1665 & 0.1697 & 3 & 20 & 0.1461 & 0.6028 \\
MoNav-TT & 0 & 7 & 0.2849 & 0.3989 & 1 & 5 & 0.0893 & 0.4863 \\
TRAOD & 4 & 7 & 0.1060& 0.5498 & 0 & 1 & 0.0030 & 0.4390 \\
DSVDD & 7 & 15 & 0.6246 & 0.7714 & 1 & 14 & 0.1010 & 0.4911 \\
DAE & 5 & 12 & 0.4627 & 0.6234 & 4 & 19 & 0.1466 & 0.5641 \\ 
GM-VSAE    & 4                           & 13                              & 0.4892                   & 0.6034  & 2           & 12           & 0.1243          & 0.5482            \\
DeepTEA     & 6                           & 14                              & 0.5290                   & 0.7540  & 4           & 22           & 0.1752          & 0.6398                \\
\midrule
Ours-MLP & \textbf{8} & \textbf{17} & \textbf{0.8512} & \textbf{0.9397} & 4 & 28 & 0.2632 & 0.6737 \\
Ours-RNN & 7 & 11 & 0.6359 & 0.7467 & 3 & 12 & 0.1310 & 0.5294 \\
Ours-CNN & 6 & 16 & 0.6756 & 0.8542 & \textbf{10} & \textbf{40} & \textbf{0.4572} & \textbf{0.7141} \\
Ours-Transformer & 7 & 16 & 0.6283 & 0.8889 & 8 & 27 & 0.2783 & 0.6852 \\
\bottomrule
\end{tabular}
    \end{adjustbox}
    \caption{Outlier detection performance for all datasets. The best performance for AP and AUC scores is highlighted for each dataset. \textsuperscript{*}We report Top-25 Hits instead of Top-100 for Geolife dataset due to its size constraint. \textsuperscript{**}For implementation of comparison methods DSVDD and DAE, we only report the best performance of deep learning based competitive methods among the choice of four deep encoders (MLP, RNN, CNN and Transformer) for each dataset due to the limitation of the space.}\label{table:performance}
    \vspace{-0.5cm}
\end{table*}

\subsubsection{Competitor Methods}\label{appendix:comarison}
We compare with several unsupervised trajectory outlier detection methods, including three rule-based non-deep learning methods and two state-of-the-art deep learning methods:\\
\textbf{OMPAD} \cite{basharat2008learning} is an outlier detection method that analyzes objects' movement patterns by counting the types of locations they visit. It identifies abnormal activities by measuring the deviations in moving trends compared to established normal patterns.\\
\textbf{MoNav-TT }\cite{zhang2012smarter} is an outlier detection algorithm tailored for urban human trajectory networks, where it detects outliers by measuring discrepancies in traffic distances. In particular, a user is identified as an outlier if the traveled distance significantly deviates from their previous behavior.\\
\textbf{TRAOD } \cite{lee2008trajectory} is a partition-and-detect
framework for trajectory outlier detection, which partitions a
trajectory into a set of line segments, and then, detects outlying
line segments for trajectory outliers.\\
\textbf{DSVDD} \cite{ruff2018deep} is a deep one-class classification based outlier detection method. We generalize it to handle the task of semantic trajectory outlier detection in a most intuitive way. We map the weekly trajectories of each user to a high dimensional sphere by a deep neural network encoder. Then the distance of trajectories from the sphere's surface is quantified as an outlier score.\\
\textbf{DAE} \cite{zhou2017anomaly,dotti2020hierarchical} is a widely-used outlier detection method that leverages a deep autoencoder. Utilizing an encoder-decoder model architecture, it reconstructs input trajectories, and the resulting reconstruction error is used as an outlier indicator, signifying deviations from the normal pattern.\\
\textbf{GM-VSAE} \cite{liu2020online} introduces a deep generative model called Gaussian Mixture Variational Sequence AutoEncoder (GM-VSAE) for anomalous trajectory detection. GM-VSAE excels in capturing complex sequential information within trajectories, representing different types of normal routes in a continuous latent space, and facilitating efficient anomaly detection.\\
\textbf{DeepTEA} \cite{han2022deeptea} is a recently proposed deep learning framework designed for time-dependent trajectory outlier detection by capturing the dynamics of traffic patterns and the temporal dependencies of movements. It uses a combination of convolutional neural networks (CNNs) and recurrent neural networks (RNNs) to learn the normal patterns of trajectories over time. This approach allows DeepTEA to effectively identify outliers by comparing new trajectory data against learned patterns, taking into account both spatial and temporal characteristics, thus providing accurate and efficient online detection of anomalous trajectories.

\begin{table*}[t]
    \centering
    \begin{adjustbox}{width=1.3\columnwidth,center}
        \begin{tabular}{lcccc|cccc}
            \hline
                   & \multicolumn{4}{c}{ATL} &        & \multicolumn{2}{c}{NOLA}                                                    \\
            \cmidrule(lr){2-5} \cmidrule(l){6-9}
                   & Hungry                  & Social & Work                     & Total    & Hungry  & Social & Work    & Total    \\
            \hline
            Red    & 5 (30)                  & 1 (10) & 10 (10)                  & 16 (50)  & 8 (30)  & 0 (10) & 10 (10) & 18 (50)  \\
            Orange & 13 (30)                 & 0 (10) & 8 (10)                   & 21 (50)  & 4 (30)  & 0 (10) & 9 (10)  & 13 (50)  \\
            Yellow & 3 (30)                  & 0 (10) & 2 (10)                   & 5 (50)   & 7 (30)  & 1 (10) & 7 (10)  & 15 (50)  \\
            Total  & 21 (90)                 & 1 (30) & 20 (30)                  & 42 (150) & 19 (90) & 1 (30) & 26 (30) & 46 (150) \\
            \hline
        \end{tabular}
    \end{adjustbox}
   
    \caption{Detailed detection Top-100 hits for different types of outliers and intensity levels (red, orange, and yellow denotes 100\%, 50\%, and 20\% abnormal behavior rate over time, respectively. The outlier number in parentheses.)}
    \label{table:detection_rate}
    \vspace{-0.5cm}
\end{table*}
\subsubsection{Evaluation Metrics}
To evaluate outlier detection performance, we employ the Top-K hits metrics, where the agents with the K highest outlier scores are classified as outliers. The number of hits reflects the method's ability to distinguish outliers. Specifically, we use Top-10 and Top-100 Hits to reflect the method's ability to distinguish outliers, which aligns with the size of our datasets. In addition, we utilize Average Precision (AP) and the area under the receiver operating characteristic curve (AUC) scores, which are widely used evaluation metrics for outlier detection tasks.

\subsubsection{Implementation Details}
\vspace{-0.1cm}
Our proposed method serves as a general framework allowing for the integration of various commonly used deep representation learning techniques on trajectory data as the encoder part. To ensure a rigorous and fair comparison with competitive deep learning methods, we adopt the same deep encoders for all methods, including multilayer perceptron (MLP) \cite{cybenko1989approximation}, recurrent neural networks (RNN) \cite{hochreiter1997long}, 1-dimensional convolutional neural networks (CNN) \cite{lecun1998gradient}, and transformer encoder \cite{vaswani2017attention}. Additionally, to ensure fairness in our comparison, all deep models adhere to a uniform architecture, characterized by each daily trajectory with a cutoff length of 16, $L=4$ encoder layers, a hidden dimension of $d=64$, 200 training epochs, and an adaptive learning rate starting from $5e^{-3}$ with a decay rate of 0.9 for every 50 training epochs. Training is executed through back-propagation using the Adam optimizer \cite{kingma2014adam}, with batch sizes of 128 for regular size datasets and 32 for the ATL-large dataset. The experimental process is conducted on four NVIDIA A100-80GB GPUs. 

\subsection{Outlier Detection Results}
\subsubsection{Main Detection Results}
The outlier detection performance of both our proposed method and competitive methods are presented in Table \ref{table:performance}. We summarize the following observations:

\noindent 1. The results demonstrate the superior outlier detection strength of our proposed contrastive learning method by consistently achieving the best performance across all datasets. It surpasses the second-best method with an average improvement of 0.148 in AUC scores and an additional 16.2 in Top-100 Hits. Notably, our approach achieves a perfect score, with 10 out of 10 hits in the top 10 outlier scores, on five of the six datasets.

\noindent 2. We observe performance variations among different encoder choices. The 1D CNN encoder delivers the best performance in five out of six datasets, which may be attributed to its simplicity and effectiveness in extracting sequential patterns. Conversely, the RNN encoder, although outperforming most competitive methods, ranks lowest among our encoders, may be explained by its well-known issue of vanishing gradients in representing long sequences.

\noindent 3. Deep learning-based methods outperform traditional ones by an average of 33.47\% in AUC scores and an additional 20.42 in Top-100 Hits. This indicates that non-deep learning methods may struggle to adequately represent complex semantic trajectories, limiting their outlier detection efficacy.

\noindent 4. There is worth noting that different encoder models (MLP, RNN, CNN and Transformer) exhibit relatively diversified performance. Especially, the Transformer's performance is nearly on par with CNN models, exhibiting only a 3.5\% average gap. This gap in performance could be attributed to the limited amount of data available, as Transformers, with their higher number of trainable parameters, generally require more data for training. Additionally, the lower efficacy of RNNs may due to the challenges in the optimization process, commonly referred to as the issue of vanishing gradients.

\subsubsection{Detailed Detection Ratio for Types of Outlier.}
Besides the superior performance, it is interesting to understand what kinds of outliers and to what degree they can be detected by our algorithm. As previously mentioned that three kinds of outliers (Hunger, Social, and Work) and three abnormal intensity levels (Red, Orange, and Yellow) exist in the simulated datasets. Here, we report the detection rate of each category for Top-100 Hits in Table \ref{table:detection_rate}. The designed model demonstrates the ability to detect most outliers of the ``Work" type but can barely detect those of the ``Social" type. This may suggest that the method at recognizing location pattern changes but is less sensitive to variations in travel distances. The detection of social outliers proves to be significantly more challenging compared to the other two categories, and detecting a YELLOW level outlier is also more challenging than the other two.


\vspace{-0.2cm}
\subsubsection{Transfer Ability Analysis.}
We continue to explore the transfer capability of our proposed contrastive learning method. In real-world scenarios, there are often situations where it would be advantageous to apply a model trained on an existing dataset to a new, unseen dataset without additional training. This approach serves two primary objectives: (1) to conserve computational resources, as training a model from scratch can be both time-intensive and resource-consuming; (2) to mitigate challenges that an unseen dataset does not contain sufficient data to train a model effectively. To evaluate the efficacy of transferring our trained model to unseen datasets, we directly apply the trained model on source datasets under four transfer situations: ATL$\rightarrow$FVA, FVA$\rightarrow$ATL, ATL$\rightarrow$NOLA, and FVA$\rightarrow$NOLA, without any further adjustments. It is noteworthy that these source and target datasets comprise different user sets and different cities. 

From the results in Table \ref{table:transfer_results}, it is evident that the transfer model, when applied from the source to the target dataset, can achieve performance on par with, or in some instances even surpassing, the model directly trained on the source dataset. Notably, when employing a CNN as the encoder, we observed even better transfer performance in three out of four cases compared to the model directly trained on the target datasets. These results demonstrate our model's effective transfer across datasets, addressing computational and data scarcity challenges. 

\begin{table}[t]
    \centering
    \begin{adjustbox}{width=1.0\columnwidth,center}
        \begin{tabular}{cccccc}
            \toprule
            Dataset      & Encoders       & Top-100 Hits & AP Score & AUC Score \\
            \midrule
            ATL          & MLP | original & 32           & 0.2509   & 0.6561    \\
            $\downarrow$ & MLP | transfer & 29           & 0.2346   & 0.6479    \\
            \cmidrule(lr){2-5}
            FVA          & CNN | original & 40           & 0.3151   & 0.6669    \\
                         & CNN | transfer & 41           & 0.3111   & 0.6596    \\
            \midrule
            FVA          & MLP | original & 32           & 0.2509   & 0.6561    \\
            $\downarrow$ & MLP | transfer & 25           & 0.2069   & 0.6345    \\
            \cmidrule(lr){2-5}
            ATL          & CNN | original & 40           & 0.3151   & 0.6669    \\
                         & CNN | transfer & 38           & 0.3051   & 0.7021    \\
            \midrule
            ATL          & MLP | original & 41           & 0.3376   & 0.6985    \\
            $\downarrow$ & MLP | transfer & 33           & 0.2461   & 0.6593    \\
            \cmidrule(lr){2-5}
            NOLA         & CNN | original & 46           & 0.3631   & 0.7185    \\
                         & CNN | transfer & 48           & 0.3718   & 0.7227    \\
            \midrule
            FVA          & MLP | original & 41           & 0.3376   & 0.6985    \\
            $\downarrow$ & MLP | transfer & 37           & 0.2945   & 0.6622    \\
            \cmidrule(lr){2-5}
            NOLA         & CNN | original & 46           & 0.3631   & 0.7185    \\
                         & CNN | transfer & 47           & 0.3657   & 0.7134    \\
            \bottomrule
        \end{tabular}
    \end{adjustbox}
    \caption{Transfer learning ability results for ATL to FVA and NOLA, and FVA to ATL and NOLA datasets. Here `original' denotes training the model on target dataset from scratch, while `transfer' denotes to apply the trained model on source dataset to target dataset without further tuning. \vspace{-0.1cm}}
    \label{table:transfer_results}
    \vspace{-0.6cm}
\end{table}

\vspace{-0.2cm}
\subsubsection{Ablation Study.}
Here, we investigate the impact of the proposed components of our method. We consider three variants of our model \textit{No-Semantic}, \textit{No-Spatial} and \textit{No-Temporal}, which remove the semantic, spatial or temporal information, separately. We report the results on ATL dataset with CNN encoder in Table~\ref{table:ablation}, where the results on other datasets with other encoders are similar. Our findings indicate that every component is vital for our method's success, with performance declining upon the removal of any part.

\vspace{-0.2cm}
\subsubsection{Parameter Sensitivity Analysis}
Here, we further conduct sensitivity analysis on the important parameters involved in our experiments. (1) We first deploy an experiment on extending the test time periods. The test period with outliers is set as two weeks in our datasets. However, it is usually difficult to determine the exact time when anomalous behavior starts. Here, we extend our test period to a longer time that includes more days before the time point anomalous behaviors start. In specific, we extend the 2 weeks test period to 4,6 and 8 weeks. We report the results on ATL dataset with CNN encoder in Table \ref{table:sensitivity1}. We can observe that the model maintains strong detection ability, with only a small drop in performance, even when extending the test period fourfold to 8 weeks. The findings demonstrate significant robustness, with a mere 7.6\% decline in performance even when the noise level is quadrupled relative to the signal, which still outperforms all comparison method when no noise present. (2) We tested the model's performance with respect to the number of training epochs, varying from 50 to 500 epochs, as shown in Table \ref{table:sensitivity1}. The results demonstrate the robustness of our model is not sensitive to parameter changes, exhibiting only a 3.16\% of variation in ROC scores.

\begin{table}[t]
    \centering
    \begin{adjustbox}{width=.99\columnwidth,center}
        \begin{tabular}{lcccc}
            \toprule
            Category             & Top-10 Hits & Top-100 Hits & AP Score & AUC Score \\
            \midrule
            Full                 & 10          & 42           & 0.3205   & 0.7215    \\
            \textit{No-Semantic} & 9           & 32           & 0.2532   & 0.6262    \\
            \textit{No-Spatial}  & 10          & 39           & 0.3017   & 0.7000    \\
            \textit{No-Temporal} & 10          & 35           & 0.2908   & 0.6897    \\
            \bottomrule
        \end{tabular}
    \end{adjustbox}
    \caption{Ablation studies. Comparison with the full model.\vspace{-0.1cm}}
    \label{table:ablation}
    \vspace{-0.5cm}
\end{table}
\begin{table}[t]
    \centering
    \begin{adjustbox}{width=.99\columnwidth,center}
        \begin{tabular}{lcc|lcc}
            \hline
            Test period & AP     & ROC    & \# epochs & AP     & ROC    \\
            \hline
            2 weeks      & 0.3205 & 0.7215 & 50        & 0.2715 & 0.6321 \\
            4 weeks      & 0.3084 & 0.7084 & 100       & 0.2698 & 0.6753 \\
            6 weeks      & 0.2768 & 0.6875 & 200       & 0.2783 & 0.6852 \\
            8 week      & 0.2433 & 0.6675 & 500       & 0.3046 & 0.6554 \\
            \hline
        \end{tabular}
    \end{adjustbox}
    \caption{Parameter sensitivity analysis on comparison of test period spans and number of epochs.\vspace{-0.1cm}}
    \label{table:sensitivity1}
    \vspace{-0.5cm}
\end{table}

\vspace{-0.2cm}
\subsubsection{Efficiency Analysis}
In Table \ref{table:running_time}, we present the running time per epoch at a range of training time spans from 1 month to 121 months with 1,000 agents. The results reveal that the running time for most encoders (MLP, CNN, and Transformer) follows an approximately linear growth trend with respect to the increase in training trajectory time span. On the other hand, the slow running time with the RNN model may be attributed to its recursive structure, which can affect efficiency in large-scale parallel computing.

\begin{table}[t]
    \centering
    \begin{adjustbox}{width=0.75\columnwidth,center}
        \begin{tabular}{lccc}
            \hline
            Method      & 1 mon & 15 mon & 121 mon \\
            \hline
            MLP         & 0.779 & 11.578 & 110.334 \\
            RNN         & 0.955 & 18.108 & 614.385 \\
            CNN         & 0.784 & 10.741 & 113.833 \\
            Transformer & 1.142 & 13.405 & 131.646 \\
            \hline
        \end{tabular}
    \end{adjustbox}
    \caption{Comparison of running time per epoch over different encoders and trajectory time spans (Unit: second). \vspace{-0.2cm}}
    \label{table:running_time}
    \vspace{-0.5cm}
\end{table}

\section{Conclusions and Future Work}
\label{sec:conclusion}

In conclusion, this study advances the domain of outlier detection in human semantic trajectories by introducing a novel self-supervised learning approach that leverages the inherent temporal periodicity in human mobility behaviors. Traditional methods, which typically relied on hand-crafted spatiotemporal indicators, have been shown to possess limitations in their adaptability to unseen outlier patterns. In contrast, our methodology, built on intuitive human behavior patterns, presents a promising solution for detecting meaningful outliers of semantic trajectories. The comprehensive experiments confirmed the effectiveness, robustness, and efficiency of our proposed method. For future research directions, we are inclined to investigate the underlying factors influencing disparate model performances across various outlier types. Additionally, refining our approach to accommodate corner cases, such as holidays, may enhance the robustness of outlier detection in real-world scenarios.



\bibliographystyle{ACM-Reference-Format}
\bibliography{main}


\begin{thebibliography}{58}


\ifx \showCODEN    \undefined \def \showCODEN     #1{\unskip}     \fi
\ifx \showDOI      \undefined \def \showDOI       #1{#1}\fi
\ifx \showISBNx    \undefined \def \showISBNx     #1{\unskip}     \fi
\ifx \showISBNxiii \undefined \def \showISBNxiii  #1{\unskip}     \fi
\ifx \showISSN     \undefined \def \showISSN      #1{\unskip}     \fi
\ifx \showLCCN     \undefined \def \showLCCN      #1{\unskip}     \fi
\ifx \shownote     \undefined \def \shownote      #1{#1}          \fi
\ifx \showarticletitle \undefined \def \showarticletitle #1{#1}   \fi
\ifx \showURL      \undefined \def \showURL       {\relax}        \fi
\providecommand\bibfield[2]{#2}
\providecommand\bibinfo[2]{#2}
\providecommand\natexlab[1]{#1}
\providecommand\showeprint[2][]{arXiv:#2}

\bibitem[Alvares et~al\mbox{.}(2007)]%
        {alvares2007towards}
\bibfield{author}{\bibinfo{person}{Luis~Otavio Alvares}, \bibinfo{person}{Vania Bogorny}, {et~al\mbox{.}}} \bibinfo{year}{2007}\natexlab{}.
\newblock \showarticletitle{Towards semantic trajectory knowledge discovery}.
\newblock \bibinfo{journal}{\emph{Data Mining and Knowledge Discovery}}  \bibinfo{volume}{12} (\bibinfo{year}{2007}).
\newblock


\bibitem[Amiri et~al\mbox{.}(2024a)]%
        {amiri2024patterns}
\bibfield{author}{\bibinfo{person}{Hossein Amiri}, \bibinfo{person}{Will Kohn}, {et~al\mbox{.}}} \bibinfo{year}{2024}\natexlab{a}.
\newblock \showarticletitle{The Patterns of Life Human Mobility Simulation}.
\newblock  (\bibinfo{year}{2024}).
\newblock
\showeprint{2410.00185}


\bibitem[Amiri et~al\mbox{.}(2024b)]%
        {amiri2024urban}
\bibfield{author}{\bibinfo{person}{Hossein Amiri}, \bibinfo{person}{Ruochen Kong}, {and} \bibinfo{person}{Andreas Zufle}.} \bibinfo{year}{2024}\natexlab{b}.
\newblock \showarticletitle{Urban Anomalies: A Simulated Human Mobility Dataset with Injected Anomalies}.
\newblock  (\bibinfo{year}{2024}).
\newblock
\showeprint{2410.01844}


\bibitem[Amiri et~al\mbox{.}(2023)]%
        {amiri2023massive}
\bibfield{author}{\bibinfo{person}{Hossein Amiri}, \bibinfo{person}{Shiyang Ruan}, {et~al\mbox{.}}} \bibinfo{year}{2023}\natexlab{}.
\newblock \showarticletitle{Massive Trajectory Data Based on Patterns of Life}. In \bibinfo{booktitle}{\emph{SIGSPATIAL'23}}. \bibinfo{publisher}{ACM}, \bibinfo{pages}{1--4}.
\newblock


\bibitem[Atwal et~al\mbox{.}(2022)]%
        {atwal2022predicting}
\bibfield{author}{\bibinfo{person}{Kuldip~Singh Atwal}, \bibinfo{person}{Taylor Anderson}, \bibinfo{person}{Dieter Pfoser}, {and} \bibinfo{person}{Andreas Z{\"u}fle}.} \bibinfo{year}{2022}\natexlab{}.
\newblock \showarticletitle{Predicting building types using OpenStreetMap}.
\newblock \bibinfo{journal}{\emph{Scientific Reports}} \bibinfo{volume}{12}, \bibinfo{number}{1} (\bibinfo{year}{2022}), \bibinfo{pages}{19976}.
\newblock


\bibitem[Basharat et~al\mbox{.}(2008)]%
        {basharat2008learning}
\bibfield{author}{\bibinfo{person}{Arslan Basharat}, \bibinfo{person}{Alexei Gritai}, {and} \bibinfo{person}{Mubarak Shah}.} \bibinfo{year}{2008}\natexlab{}.
\newblock \showarticletitle{Learning object motion patterns for anomaly detection and improved object detection}. In \bibinfo{booktitle}{\emph{2008 IEEE conference on computer vision and pattern recognition}}. IEEE, \bibinfo{pages}{1--8}.
\newblock


\bibitem[Belhadi et~al\mbox{.}(2020)]%
        {Belhadi2020}
\bibfield{author}{\bibinfo{person}{Asma Belhadi}, \bibinfo{person}{Youcef Djenouri}, \bibinfo{person}{Jerry Chun-Wei Lin}, {and} \bibinfo{person}{Alberto Cano}.} \bibinfo{year}{2020}\natexlab{}.
\newblock \showarticletitle{Trajectory Outlier Detection: Algorithms, Taxonomies, Evaluation, and Open Challenges}.
\newblock \bibinfo{journal}{\emph{ACM Trans. Manage. Inf. Syst.}} \bibinfo{volume}{11}, \bibinfo{number}{3}, Article \bibinfo{articleno}{16} (\bibinfo{date}{jun} \bibinfo{year}{2020}), \bibinfo{numpages}{29}~pages.
\newblock
\showISSN{2158-656X}
\urldef\tempurl%
\url{https://doi.org/10.1145/3399631}
\showDOI{\tempurl}


\bibitem[Bennett(2010)]%
        {bennett2010openstreetmap}
\bibfield{author}{\bibinfo{person}{Jonathan Bennett}.} \bibinfo{year}{2010}\natexlab{}.
\newblock \bibinfo{booktitle}{\emph{OpenStreetMap}}.
\newblock \bibinfo{publisher}{Packt Publishing Ltd}.
\newblock


\bibitem[Chen et~al\mbox{.}(2020b)]%
        {chen2020parallel}
\bibfield{author}{\bibinfo{person}{Lisi Chen}, \bibinfo{person}{Shuo Shang}, \bibinfo{person}{Christian~S Jensen}, \bibinfo{person}{Bin Yao}, {and} \bibinfo{person}{Panos Kalnis}.} \bibinfo{year}{2020}\natexlab{b}.
\newblock \showarticletitle{Parallel semantic trajectory similarity join}. In \bibinfo{booktitle}{\emph{2020 IEEE 36th International Conference on Data Engineering (ICDE)}}. IEEE, \bibinfo{pages}{997--1008}.
\newblock


\bibitem[Chen et~al\mbox{.}(2020a)]%
        {chen2020simple}
\bibfield{author}{\bibinfo{person}{Ting Chen}, \bibinfo{person}{Simon Kornblith}, \bibinfo{person}{Mohammad Norouzi}, {and} \bibinfo{person}{Geoffrey Hinton}.} \bibinfo{year}{2020}\natexlab{a}.
\newblock \showarticletitle{A simple framework for contrastive learning of visual representations}. In \bibinfo{booktitle}{\emph{International conference on machine learning}}. PMLR, \bibinfo{pages}{1597--1607}.
\newblock


\bibitem[Cong et~al\mbox{.}(2012)]%
        {cong2012efficient}
\bibfield{author}{\bibinfo{person}{Gao Cong}, \bibinfo{person}{Hua Lu}, \bibinfo{person}{Beng~Chin Ooi}, \bibinfo{person}{Dongxiang Zhang}, {and} \bibinfo{person}{Meihui Zhang}.} \bibinfo{year}{2012}\natexlab{}.
\newblock \showarticletitle{Efficient spatial keyword search in trajectory databases}.
\newblock \bibinfo{journal}{\emph{arXiv preprint arXiv:1205.2880}} (\bibinfo{year}{2012}).
\newblock


\bibitem[Cybenko(1989)]%
        {cybenko1989approximation}
\bibfield{author}{\bibinfo{person}{George Cybenko}.} \bibinfo{year}{1989}\natexlab{}.
\newblock \showarticletitle{Approximation by superpositions of a sigmoidal function}.
\newblock \bibinfo{journal}{\emph{Mathematics of control, signals and systems}} \bibinfo{volume}{2}, \bibinfo{number}{4} (\bibinfo{year}{1989}), \bibinfo{pages}{303--314}.
\newblock


\bibitem[Daneshpazhouh and Sami(2014)]%
        {Daneshpazhouh2014}
\bibfield{author}{\bibinfo{person}{Armin Daneshpazhouh} {and} \bibinfo{person}{Ashkan Sami}.} \bibinfo{year}{2014}\natexlab{}.
\newblock \showarticletitle{{Entropy-based outlier detection using semi-supervised approach with few positive examples}}.
\newblock \bibinfo{journal}{\emph{Pattern Recognition Letters}}  \bibinfo{volume}{49} (\bibinfo{date}{nov} \bibinfo{year}{2014}), \bibinfo{pages}{77--84}.
\newblock
\showISSN{01678655}
\urldef\tempurl%
\url{https://doi.org/10.1016/j.patrec.2014.06.012}
\showDOI{\tempurl}


\bibitem[Dotti et~al\mbox{.}(2020)]%
        {dotti2020hierarchical}
\bibfield{author}{\bibinfo{person}{Dario Dotti}, \bibinfo{person}{Mirela Popa}, {and} \bibinfo{person}{Stylianos Asteriadis}.} \bibinfo{year}{2020}\natexlab{}.
\newblock \showarticletitle{A hierarchical autoencoder learning model for path prediction and abnormality detection}.
\newblock \bibinfo{journal}{\emph{Pattern Recognition Letters}}  \bibinfo{volume}{130} (\bibinfo{year}{2020}), \bibinfo{pages}{216--224}.
\newblock


\bibitem[Gonzalez et~al\mbox{.}(2008)]%
        {gonzalez2008understanding}
\bibfield{author}{\bibinfo{person}{Marta~C Gonzalez}, \bibinfo{person}{Cesar~A Hidalgo}, {and} \bibinfo{person}{Albert-Laszlo Barabasi}.} \bibinfo{year}{2008}\natexlab{}.
\newblock \showarticletitle{Understanding individual human mobility patterns}.
\newblock \bibinfo{journal}{\emph{nature}} \bibinfo{volume}{453}, \bibinfo{number}{7196} (\bibinfo{year}{2008}), \bibinfo{pages}{779--782}.
\newblock


\bibitem[Gupta et~al\mbox{.}(2014)]%
        {Gupta2014}
\bibfield{author}{\bibinfo{person}{Manish Gupta}, \bibinfo{person}{Jing Gao}, \bibinfo{person}{Charu~C. Aggarwal}, {and} \bibinfo{person}{Jiawei Han}.} \bibinfo{year}{2014}\natexlab{}.
\newblock \showarticletitle{{Outlier Detection for Temporal Data: A Survey}}.
\newblock \bibinfo{journal}{\emph{IEEE Transactions on Knowledge and Data Engineering}} \bibinfo{volume}{26}, \bibinfo{number}{9} (\bibinfo{date}{sep} \bibinfo{year}{2014}), \bibinfo{pages}{2250--2267}.
\newblock
\showISSN{1041-4347}
\urldef\tempurl%
\url{https://doi.org/10.1109/TKDE.2013.184}
\showDOI{\tempurl}


\bibitem[Hadsell et~al\mbox{.}(2006)]%
        {hadsell2006dimensionality}
\bibfield{author}{\bibinfo{person}{Raia Hadsell}, \bibinfo{person}{Sumit Chopra}, {and} \bibinfo{person}{Yann LeCun}.} \bibinfo{year}{2006}\natexlab{}.
\newblock \showarticletitle{Dimensionality reduction by learning an invariant mapping}. In \bibinfo{booktitle}{\emph{2006 IEEE Computer Society Conference on Computer Vision and Pattern Recognition (CVPR'06)}}, Vol.~\bibinfo{volume}{2}. IEEE, \bibinfo{pages}{1735--1742}.
\newblock


\bibitem[Han et~al\mbox{.}(2022)]%
        {han2022deeptea}
\bibfield{author}{\bibinfo{person}{Xiaolin Han}, \bibinfo{person}{Reynold Cheng}, \bibinfo{person}{Chenhao Ma}, {and} \bibinfo{person}{Tobias Grubenmann}.} \bibinfo{year}{2022}\natexlab{}.
\newblock \showarticletitle{DeepTEA: effective and efficient online time-dependent trajectory outlier detection}.
\newblock \bibinfo{journal}{\emph{Proceedings of the VLDB Endowment}} \bibinfo{volume}{15}, \bibinfo{number}{7} (\bibinfo{year}{2022}), \bibinfo{pages}{1493--1505}.
\newblock


\bibitem[He et~al\mbox{.}(2020)]%
        {he2020momentum}
\bibfield{author}{\bibinfo{person}{Kaiming He}, \bibinfo{person}{Haoqi Fan}, \bibinfo{person}{Yuxin Wu}, \bibinfo{person}{Saining Xie}, {and} \bibinfo{person}{Ross Girshick}.} \bibinfo{year}{2020}\natexlab{}.
\newblock \showarticletitle{Momentum contrast for unsupervised visual representation learning}. In \bibinfo{booktitle}{\emph{Proceedings of the IEEE/CVF conference on computer vision and pattern recognition}}. \bibinfo{pages}{9729--9738}.
\newblock


\bibitem[Hochreiter and Schmidhuber(1997)]%
        {hochreiter1997long}
\bibfield{author}{\bibinfo{person}{Sepp Hochreiter} {and} \bibinfo{person}{J{\"u}rgen Schmidhuber}.} \bibinfo{year}{1997}\natexlab{}.
\newblock \showarticletitle{Long short-term memory}.
\newblock \bibinfo{journal}{\emph{Neural computation}} \bibinfo{volume}{9}, \bibinfo{number}{8} (\bibinfo{year}{1997}), \bibinfo{pages}{1735--1780}.
\newblock


\bibitem[Kim et~al\mbox{.}(2020)]%
        {kim2020location}
\bibfield{author}{\bibinfo{person}{Joon-Seok Kim}, \bibinfo{person}{Hyunjee Jin}, {et~al\mbox{.}}} \bibinfo{year}{2020}\natexlab{}.
\newblock \showarticletitle{Location-based social network data generation based on patterns of life}. In \bibinfo{booktitle}{\emph{2020 21st IEEE International Conference on Mobile Data Management (MDM)}}. IEEE, \bibinfo{pages}{158--167}.
\newblock


\bibitem[Kingma and Ba(2014)]%
        {kingma2014adam}
\bibfield{author}{\bibinfo{person}{Diederik~P Kingma} {and} \bibinfo{person}{Jimmy Ba}.} \bibinfo{year}{2014}\natexlab{}.
\newblock \showarticletitle{Adam: A method for stochastic optimization}.
\newblock \bibinfo{journal}{\emph{arXiv preprint arXiv:1412.6980}} (\bibinfo{year}{2014}).
\newblock


\bibitem[Kohn et~al\mbox{.}(2023)]%
        {kohn2023epipol}
\bibfield{author}{\bibinfo{person}{Will Kohn}, \bibinfo{person}{Hossein Amiri}, {and} \bibinfo{person}{Andreas Z{\"u}fle}.} \bibinfo{year}{2023}\natexlab{}.
\newblock \showarticletitle{EPIPOL: An Epidemiological Patterns of Life Simulation (Demonstration Paper)}. In \bibinfo{booktitle}{\emph{SIGSPATIAL SpatialEpi'23 Workshop}}. \bibinfo{publisher}{ACM}, \bibinfo{pages}{13--16}.
\newblock


\bibitem[LeCun et~al\mbox{.}(1998)]%
        {lecun1998gradient}
\bibfield{author}{\bibinfo{person}{Yann LeCun}, \bibinfo{person}{L{\'e}on Bottou}, \bibinfo{person}{Yoshua Bengio}, {and} \bibinfo{person}{Patrick Haffner}.} \bibinfo{year}{1998}\natexlab{}.
\newblock \showarticletitle{Gradient-based learning applied to document recognition}.
\newblock \bibinfo{journal}{\emph{Proc. IEEE}} \bibinfo{volume}{86}, \bibinfo{number}{11} (\bibinfo{year}{1998}), \bibinfo{pages}{2278--2324}.
\newblock


\bibitem[Lee et~al\mbox{.}(2008)]%
        {lee2008trajectory}
\bibfield{author}{\bibinfo{person}{Jae-Gil Lee}, \bibinfo{person}{Jiawei Han}, {and} \bibinfo{person}{Xiaolei Li}.} \bibinfo{year}{2008}\natexlab{}.
\newblock \showarticletitle{Trajectory outlier detection: A partition-and-detect framework}. In \bibinfo{booktitle}{\emph{2008 IEEE 24th International Conference on Data Engineering}}. IEEE, \bibinfo{pages}{140--149}.
\newblock


\bibitem[Leskovec and Sosi{\v{c}}(2016)]%
        {leskovec2016snap}
\bibfield{author}{\bibinfo{person}{Jure Leskovec} {and} \bibinfo{person}{Rok Sosi{\v{c}}}.} \bibinfo{year}{2016}\natexlab{}.
\newblock \showarticletitle{SNAP: A General-Purpose Network Analysis and Graph-Mining Library}.
\newblock \bibinfo{journal}{\emph{ACM Transactions on Intelligent Systems and Technology (TIST)}} \bibinfo{volume}{8}, \bibinfo{number}{1} (\bibinfo{year}{2016}), \bibinfo{pages}{1}.
\newblock


\bibitem[Liu and Guo(2020)]%
        {liu2020stccd}
\bibfield{author}{\bibinfo{person}{Caihong Liu} {and} \bibinfo{person}{Chonghui Guo}.} \bibinfo{year}{2020}\natexlab{}.
\newblock \showarticletitle{STCCD: Semantic trajectory clustering based on community detection in networks}.
\newblock \bibinfo{journal}{\emph{Expert Systems with Applications}}  \bibinfo{volume}{162} (\bibinfo{year}{2020}), \bibinfo{pages}{113689}.
\newblock


\bibitem[Liu et~al\mbox{.}(2013)]%
        {liu2013moir}
\bibfield{author}{\bibinfo{person}{Kuien Liu}, \bibinfo{person}{Bin Yang}, \bibinfo{person}{Shuo Shang}, \bibinfo{person}{Yaguang Li}, {and} \bibinfo{person}{Zhiming Ding}.} \bibinfo{year}{2013}\natexlab{}.
\newblock \showarticletitle{MOIR/UOTS: trip recommendation with user oriented trajectory search}. In \bibinfo{booktitle}{\emph{2013 IEEE 14th International Conference on Mobile Data Management}}, Vol.~\bibinfo{volume}{1}. IEEE, \bibinfo{pages}{335--337}.
\newblock


\bibitem[Liu et~al\mbox{.}(2024)]%
        {liu2024neural}
\bibfield{author}{\bibinfo{person}{Yueyang Liu}, \bibinfo{person}{Lance Kennedy}, \bibinfo{person}{Hossein Amiri}, {and} \bibinfo{person}{Andreas Z{\"u}fle}.} \bibinfo{year}{2024}\natexlab{}.
\newblock \bibinfo{title}{Neural Collaborative Filtering to Detect Anomalies in Human Semantic Trajectories}.
\newblock
\newblock
\showeprint{2409.18427}


\bibitem[Liu et~al\mbox{.}(2020)]%
        {liu2020online}
\bibfield{author}{\bibinfo{person}{Yiding Liu}, \bibinfo{person}{Kaiqi Zhao}, \bibinfo{person}{Gao Cong}, {and} \bibinfo{person}{Zhifeng Bao}.} \bibinfo{year}{2020}\natexlab{}.
\newblock \showarticletitle{Online anomalous trajectory detection with deep generative sequence modeling}. In \bibinfo{booktitle}{\emph{2020 IEEE 36th International Conference on Data Engineering (ICDE)}}. IEEE, \bibinfo{pages}{949--960}.
\newblock


\bibitem[Meng et~al\mbox{.}(2019)]%
        {Meng2019}
\bibfield{author}{\bibinfo{person}{Fanrong Meng}, \bibinfo{person}{Guan Yuan}, \bibinfo{person}{Shaoqian Lv}, \bibinfo{person}{Zhixiao Wang}, {and} \bibinfo{person}{Shixiong Xia}.} \bibinfo{year}{2019}\natexlab{}.
\newblock \showarticletitle{{An overview on trajectory outlier detection}}.
\newblock \bibinfo{journal}{\emph{Artificial Intelligence Review}} \bibinfo{volume}{52}, \bibinfo{number}{4} (\bibinfo{date}{dec} \bibinfo{year}{2019}), \bibinfo{pages}{2437--2456}.
\newblock
\showISSN{0269-2821}
\urldef\tempurl%
\url{https://doi.org/10.1007/s10462-018-9619-1}
\showDOI{\tempurl}


\bibitem[Mokbel et~al\mbox{.}(2020)]%
        {mokbel2020contact}
\bibfield{author}{\bibinfo{person}{Mohamed Mokbel}, \bibinfo{person}{Sofiane Abbar}, {and} \bibinfo{person}{Rade Stanojevic}.} \bibinfo{year}{2020}\natexlab{}.
\newblock \showarticletitle{Contact tracing: Beyond the apps}.
\newblock \bibinfo{journal}{\emph{SIGSPATIAL Special}} \bibinfo{volume}{12}, \bibinfo{number}{2} (\bibinfo{year}{2020}), \bibinfo{pages}{15--24}.
\newblock


\bibitem[Mokbel et~al\mbox{.}(2022)]%
        {mokbel2022mobility}
\bibfield{author}{\bibinfo{person}{Mohamed Mokbel}, \bibinfo{person}{Mahmoud Sakr}, \bibinfo{person}{Li Xiong}, \bibinfo{person}{Andreas Z{\"u}fle}, \bibinfo{person}{Jussara Almeida}, \bibinfo{person}{Taylor Anderson}, \bibinfo{person}{Walid Aref}, \bibinfo{person}{Gennady Andrienko}, \bibinfo{person}{Natalia Andrienko}, \bibinfo{person}{Yang Cao}, {et~al\mbox{.}}} \bibinfo{year}{2022}\natexlab{}.
\newblock \showarticletitle{Mobility data science (dagstuhl seminar 22021)}. In \bibinfo{booktitle}{\emph{Dagstuhl reports}}, Vol.~\bibinfo{volume}{12}. Schloss Dagstuhl-Leibniz-Zentrum f{\"u}r Informatik.
\newblock


\bibitem[Mokbel et~al\mbox{.}(2023)]%
        {mokbel2023towards}
\bibfield{author}{\bibinfo{person}{Mohamed Mokbel}, \bibinfo{person}{Mahmoud Sakr}, \bibinfo{person}{Li Xiong}, \bibinfo{person}{Andreas Z{\"u}fle}, \bibinfo{person}{Jussara Almeida}, \bibinfo{person}{Walid Aref}, \bibinfo{person}{Gennady Andrienko}, \bibinfo{person}{Natalia Andrienko}, \bibinfo{person}{Yang Cao}, \bibinfo{person}{Sanjay Chawla}, {et~al\mbox{.}}} \bibinfo{year}{2023}\natexlab{}.
\newblock \showarticletitle{Towards Mobility Data Science (Vision Paper)}.
\newblock \bibinfo{journal}{\emph{arXiv preprint arXiv:2307.05717}} (\bibinfo{year}{2023}).
\newblock


\bibitem[Oord et~al\mbox{.}(2018)]%
        {oord2018representation}
\bibfield{author}{\bibinfo{person}{Aaron van~den Oord}, \bibinfo{person}{Yazhe Li}, {and} \bibinfo{person}{Oriol Vinyals}.} \bibinfo{year}{2018}\natexlab{}.
\newblock \showarticletitle{Representation learning with contrastive predictive coding}.
\newblock \bibinfo{journal}{\emph{arXiv preprint arXiv:1807.03748}} (\bibinfo{year}{2018}).
\newblock


\bibitem[Parent et~al\mbox{.}(2013)]%
        {parent2013semantic}
\bibfield{author}{\bibinfo{person}{Christine Parent}, \bibinfo{person}{Stefano Spaccapietra}, \bibinfo{person}{Chiara Renso}, \bibinfo{person}{Gennady Andrienko}, \bibinfo{person}{Natalia Andrienko}, \bibinfo{person}{Vania Bogorny}, \bibinfo{person}{Maria~Luisa Damiani}, \bibinfo{person}{Aris Gkoulalas-Divanis}, \bibinfo{person}{Jose Macedo}, \bibinfo{person}{Nikos Pelekis}, {et~al\mbox{.}}} \bibinfo{year}{2013}\natexlab{}.
\newblock \showarticletitle{Semantic trajectories modeling and analysis}.
\newblock \bibinfo{journal}{\emph{ACM Computing Surveys (CSUR)}} \bibinfo{volume}{45}, \bibinfo{number}{4} (\bibinfo{year}{2013}), \bibinfo{pages}{1--32}.
\newblock


\bibitem[Rambhatla et~al\mbox{.}(2022)]%
        {RambhatlaZSSL22}
\bibfield{author}{\bibinfo{person}{Sirisha Rambhatla}, \bibinfo{person}{Sepanta Zeighami}, \bibinfo{person}{Kameron Shahabi}, \bibinfo{person}{Cyrus Shahabi}, {and} \bibinfo{person}{Yan Liu}.} \bibinfo{year}{2022}\natexlab{}.
\newblock \showarticletitle{Toward Accurate Spatiotemporal {COVID-19} Risk Scores Using High-Resolution Real-World Mobility Data}.
\newblock \bibinfo{journal}{\emph{{ACM} Trans. Spatial Algorithms Syst.}} \bibinfo{volume}{8}, \bibinfo{number}{2} (\bibinfo{year}{2022}), \bibinfo{pages}{1--30}.
\newblock
\urldef\tempurl%
\url{https://doi.org/10.1145/3481044}
\showDOI{\tempurl}


\bibitem[Ruff et~al\mbox{.}(2018)]%
        {ruff2018deep}
\bibfield{author}{\bibinfo{person}{Lukas Ruff}, \bibinfo{person}{Robert Vandermeulen}, \bibinfo{person}{Nico Goernitz}, \bibinfo{person}{Lucas Deecke}, \bibinfo{person}{Shoaib~Ahmed Siddiqui}, \bibinfo{person}{Alexander Binder}, \bibinfo{person}{Emmanuel M{\"u}ller}, {and} \bibinfo{person}{Marius Kloft}.} \bibinfo{year}{2018}\natexlab{}.
\newblock \showarticletitle{Deep one-class classification}. In \bibinfo{booktitle}{\emph{International conference on machine learning}}. PMLR, \bibinfo{pages}{4393--4402}.
\newblock


\bibitem[Shahid et~al\mbox{.}(2015)]%
        {Shahid2015}
\bibfield{author}{\bibinfo{person}{Nauman Shahid}, \bibinfo{person}{Ijaz~Haider Naqvi}, {and} \bibinfo{person}{Saad~Bin Qaisar}.} \bibinfo{year}{2015}\natexlab{}.
\newblock \showarticletitle{{Characteristics and classification of outlier detection techniques for wireless sensor networks in harsh environments: a survey}}.
\newblock \bibinfo{journal}{\emph{Artificial Intelligence Review}} \bibinfo{volume}{43}, \bibinfo{number}{2} (\bibinfo{date}{feb} \bibinfo{year}{2015}), \bibinfo{pages}{193--228}.
\newblock
\showISSN{0269-2821}
\urldef\tempurl%
\url{https://doi.org/10.1007/s10462-012-9370-y}
\showDOI{\tempurl}


\bibitem[Shang et~al\mbox{.}(2012)]%
        {shang2012user}
\bibfield{author}{\bibinfo{person}{Shuo Shang}, \bibinfo{person}{Ruogu Ding}, \bibinfo{person}{Bo Yuan}, \bibinfo{person}{Kexin Xie}, \bibinfo{person}{Kai Zheng}, {and} \bibinfo{person}{Panos Kalnis}.} \bibinfo{year}{2012}\natexlab{}.
\newblock \showarticletitle{User oriented trajectory search for trip recommendation}. In \bibinfo{booktitle}{\emph{Proceedings of the 15th international conference on extending database technology}}. \bibinfo{pages}{156--167}.
\newblock


\bibitem[Shi et~al\mbox{.}(2023)]%
        {Shi2023}
\bibfield{author}{\bibinfo{person}{Juntian Shi}, \bibinfo{person}{Zhicheng Pan}, \bibinfo{person}{Junhua Fang}, {and} \bibinfo{person}{Pingfu Chao}.} \bibinfo{year}{2023}\natexlab{}.
\newblock \showarticletitle{{RUTOD: real-time urban traffic outlier detection on streaming trajectory}}.
\newblock \bibinfo{journal}{\emph{Neural Computing and Applications}} \bibinfo{volume}{35}, \bibinfo{number}{5} (\bibinfo{date}{feb} \bibinfo{year}{2023}), \bibinfo{pages}{3625--3637}.
\newblock
\showISSN{0941-0643}
\urldef\tempurl%
\url{https://doi.org/10.1007/s00521-021-06294-y}
\showDOI{\tempurl}


\bibitem[Stavropoulos et~al\mbox{.}(2020)]%
        {stavropoulos2020iot}
\bibfield{author}{\bibinfo{person}{Thanos~G Stavropoulos}, \bibinfo{person}{Asterios Papastergiou}, \bibinfo{person}{Lampros Mpaltadoros}, \bibinfo{person}{Spiros Nikolopoulos}, {and} \bibinfo{person}{Ioannis Kompatsiaris}.} \bibinfo{year}{2020}\natexlab{}.
\newblock \showarticletitle{IoT wearable sensors and devices in elderly care: A literature review}.
\newblock \bibinfo{journal}{\emph{Sensors}} \bibinfo{volume}{20}, \bibinfo{number}{10} (\bibinfo{year}{2020}), \bibinfo{pages}{2826}.
\newblock


\bibitem[Su et~al\mbox{.}(2023)]%
        {Su2023}
\bibfield{author}{\bibinfo{person}{Yueyang Su}, \bibinfo{person}{Di Yao}, {and} \bibinfo{person}{Jingping Bi}.} \bibinfo{year}{2023}\natexlab{}.
\newblock \showarticletitle{{Transfer learning for region-wide trajectory outlier detection}}.
\newblock \bibinfo{journal}{\emph{IEEE Access}} (\bibinfo{year}{2023}), \bibinfo{pages}{1--1}.
\newblock
\showISSN{2169-3536}
\urldef\tempurl%
\url{https://doi.org/10.1109/ACCESS.2023.3294689}
\showDOI{\tempurl}


\bibitem[Tolea et~al\mbox{.}(2016)]%
        {tolea2016trajectory}
\bibfield{author}{\bibinfo{person}{Magdalena~I Tolea}, \bibinfo{person}{John~C Morris}, {and} \bibinfo{person}{James~E Galvin}.} \bibinfo{year}{2016}\natexlab{}.
\newblock \showarticletitle{Trajectory of mobility decline by type of dementia}.
\newblock \bibinfo{journal}{\emph{Alzheimer disease and associated disorders}} \bibinfo{volume}{30}, \bibinfo{number}{1} (\bibinfo{year}{2016}), \bibinfo{pages}{60}.
\newblock


\bibitem[Vaswani et~al\mbox{.}(2017)]%
        {vaswani2017attention}
\bibfield{author}{\bibinfo{person}{Ashish Vaswani}, \bibinfo{person}{Noam Shazeer}, \bibinfo{person}{Niki Parmar}, \bibinfo{person}{Jakob Uszkoreit}, \bibinfo{person}{Llion Jones}, \bibinfo{person}{Aidan~N Gomez}, \bibinfo{person}{{\L}ukasz Kaiser}, {and} \bibinfo{person}{Illia Polosukhin}.} \bibinfo{year}{2017}\natexlab{}.
\newblock \showarticletitle{Attention is all you need}.
\newblock \bibinfo{journal}{\emph{Advances in neural information processing systems}}  \bibinfo{volume}{30} (\bibinfo{year}{2017}).
\newblock


\bibitem[Wang et~al\mbox{.}(2020)]%
        {Wang2020}
\bibfield{author}{\bibinfo{person}{Jingwei Wang}, \bibinfo{person}{Yun Yuan}, \bibinfo{person}{Tianle Ni}, \bibinfo{person}{Yunlong Ma}, \bibinfo{person}{Min Liu}, \bibinfo{person}{Gaowei Xu}, {and} \bibinfo{person}{Weiming Shen}.} \bibinfo{year}{2020}\natexlab{}.
\newblock \showarticletitle{{Anomalous Trajectory Detection and Classification Based on Difference and Intersection Set Distance}}.
\newblock \bibinfo{journal}{\emph{IEEE Transactions on Vehicular Technology}} \bibinfo{volume}{69}, \bibinfo{number}{3} (\bibinfo{date}{mar} \bibinfo{year}{2020}), \bibinfo{pages}{2487--2500}.
\newblock
\showISSN{0018-9545}
\urldef\tempurl%
\url{https://doi.org/10.1109/TVT.2020.2967865}
\showDOI{\tempurl}


\bibitem[Yao et~al\mbox{.}(2017)]%
        {yao2017serm}
\bibfield{author}{\bibinfo{person}{Di Yao}, \bibinfo{person}{Chao Zhang}, \bibinfo{person}{Jianhui Huang}, {and} \bibinfo{person}{Jingping Bi}.} \bibinfo{year}{2017}\natexlab{}.
\newblock \showarticletitle{Serm: A recurrent model for next location prediction in semantic trajectories}. In \bibinfo{booktitle}{\emph{Proceedings of the 2017 ACM on Conference on Information and Knowledge Management}}. \bibinfo{pages}{2411--2414}.
\newblock


\bibitem[Ying et~al\mbox{.}(2011)]%
        {ying2011semantic}
\bibfield{author}{\bibinfo{person}{Josh Jia-Ching Ying}, \bibinfo{person}{Wang-Chien Lee}, \bibinfo{person}{Tz-Chiao Weng}, {and} \bibinfo{person}{Vincent~S Tseng}.} \bibinfo{year}{2011}\natexlab{}.
\newblock \showarticletitle{Semantic trajectory mining for location prediction}. In \bibinfo{booktitle}{\emph{Proceedings of the 19th ACM SIGSPATIAL international conference on advances in geographic information systems}}. \bibinfo{pages}{34--43}.
\newblock


\bibitem[Zhang(2012)]%
        {zhang2012smarter}
\bibfield{author}{\bibinfo{person}{Jianting Zhang}.} \bibinfo{year}{2012}\natexlab{}.
\newblock \showarticletitle{Smarter outlier detection and deeper understanding of large-scale taxi trip records: a case study of NYC}. In \bibinfo{booktitle}{\emph{Proceedings of the ACM SIGKDD International Workshop on Urban Computing}}. \bibinfo{pages}{157--162}.
\newblock


\bibitem[Zhang et~al\mbox{.}(2023)]%
        {zhang2023large}
\bibfield{author}{\bibinfo{person}{Zheng Zhang}, \bibinfo{person}{Hossein Amiri}, {et~al\mbox{.}}} \bibinfo{year}{2023}\natexlab{}.
\newblock \showarticletitle{Large Language Models for Spatial Trajectory Patterns Mining}.
\newblock  (\bibinfo{year}{2023}).
\newblock
\showeprint{2310.04942}


\bibitem[Zhang and Zhao(2022)]%
        {zhang2022unsupervised}
\bibfield{author}{\bibinfo{person}{Zheng Zhang} {and} \bibinfo{person}{Liang Zhao}.} \bibinfo{year}{2022}\natexlab{}.
\newblock \showarticletitle{Unsupervised deep subgraph anomaly detection}. In \bibinfo{booktitle}{\emph{2022 IEEE International Conference on Data Mining (ICDM)}}. IEEE, \bibinfo{pages}{753--762}.
\newblock


\bibitem[Zheng et~al\mbox{.}(2015)]%
        {zheng2015approximate}
\bibfield{author}{\bibinfo{person}{Bolong Zheng}, \bibinfo{person}{Nicholas~Jing Yuan}, \bibinfo{person}{Kai Zheng}, \bibinfo{person}{Xing Xie}, \bibinfo{person}{Shazia Sadiq}, {and} \bibinfo{person}{Xiaofang Zhou}.} \bibinfo{year}{2015}\natexlab{}.
\newblock \showarticletitle{Approximate keyword search in semantic trajectory database}. In \bibinfo{booktitle}{\emph{2015 IEEE 31st International Conference on Data Engineering}}. IEEE, \bibinfo{pages}{975--986}.
\newblock


\bibitem[Zheng et~al\mbox{.}(2017)]%
        {zheng2017popularity}
\bibfield{author}{\bibinfo{person}{Kai Zheng}, \bibinfo{person}{Bolong Zheng}, \bibinfo{person}{Jiajie Xu}, \bibinfo{person}{Guanfeng Liu}, \bibinfo{person}{An Liu}, {and} \bibinfo{person}{Zhixu Li}.} \bibinfo{year}{2017}\natexlab{}.
\newblock \showarticletitle{Popularity-aware spatial keyword search on activity trajectories}.
\newblock \bibinfo{journal}{\emph{World Wide Web}}  \bibinfo{volume}{20} (\bibinfo{year}{2017}), \bibinfo{pages}{749--773}.
\newblock


\bibitem[Zheng et~al\mbox{.}(2008)]%
        {zheng2008mining}
\bibfield{author}{\bibinfo{person}{Yu Zheng}, \bibinfo{person}{Xing Xie}, \bibinfo{person}{Quannan Li}, {and} \bibinfo{person}{Wei-Ying Ma}.} \bibinfo{year}{2008}\natexlab{}.
\newblock \showarticletitle{Mining user similarity based on location history}. In \bibinfo{booktitle}{\emph{Proceedings of the 16th ACM SIGSPATIAL conference on Advance in Geographical Information Systems} (\bibinfo{edition}{proceedings of the 16th acm sigspatial conference on advance in geographical information systems} ed.)}.
\newblock
\urldef\tempurl%
\url{https://www.microsoft.com/en-us/research/publication/mining-user-similarity-based-on-location-history/}
\showURL{%
\tempurl}


\bibitem[Zheng et~al\mbox{.}(2010)]%
        {zheng2010geolife}
\bibfield{author}{\bibinfo{person}{Yu Zheng}, \bibinfo{person}{Xing Xie}, \bibinfo{person}{Wei-Ying Ma}, {et~al\mbox{.}}} \bibinfo{year}{2010}\natexlab{}.
\newblock \showarticletitle{GeoLife: A collaborative social networking service among user, location and trajectory.}
\newblock \bibinfo{journal}{\emph{IEEE Data Eng. Bull.}} \bibinfo{volume}{33}, \bibinfo{number}{2} (\bibinfo{year}{2010}), \bibinfo{pages}{32--39}.
\newblock


\bibitem[Zhou and Paffenroth(2017)]%
        {zhou2017anomaly}
\bibfield{author}{\bibinfo{person}{Chong Zhou} {and} \bibinfo{person}{Randy~C Paffenroth}.} \bibinfo{year}{2017}\natexlab{}.
\newblock \showarticletitle{Anomaly detection with robust deep autoencoders}. In \bibinfo{booktitle}{\emph{Proceedings of the 23rd ACM SIGKDD international conference on knowledge discovery and data mining}}. \bibinfo{pages}{665--674}.
\newblock


\bibitem[Z{\"u}fle et~al\mbox{.}(2024)]%
        {zufle2024silico}
\bibfield{author}{\bibinfo{person}{Andreas Z{\"u}fle}, \bibinfo{person}{Dieter Pfoser}, \bibinfo{person}{Carola Wenk}, \bibinfo{person}{Andrew Crooks}, \bibinfo{person}{Hamdi Kavak}, \bibinfo{person}{Taylor Anderson}, \bibinfo{person}{Joon-Seok Kim}, \bibinfo{person}{Nathan Holt}, {and} \bibinfo{person}{Andrew Diantonio}.} \bibinfo{year}{2024}\natexlab{}.
\newblock \showarticletitle{In Silico Human Mobility Data Science: Leveraging Massive Simulated Mobility Data (Vision Paper)}.
\newblock \bibinfo{journal}{\emph{ACM Transactions on Spatial Algorithms and Systems}} \bibinfo{volume}{10}, \bibinfo{number}{2} (\bibinfo{year}{2024}), \bibinfo{pages}{1--27}.
\newblock


\bibitem[Z{\"u}fle et~al\mbox{.}(2023)]%
        {zufle2023urban}
\bibfield{author}{\bibinfo{person}{Andreas Z{\"u}fle}, \bibinfo{person}{Carola Wenk}, \bibinfo{person}{Dieter Pfoser}, \bibinfo{person}{Andrew Crooks}, \bibinfo{person}{Joon-Seok Kim}, \bibinfo{person}{Hamdi Kavak}, \bibinfo{person}{Umar Manzoor}, {and} \bibinfo{person}{Hyunjee Jin}.} \bibinfo{year}{2023}\natexlab{}.
\newblock \showarticletitle{Urban life: a model of people and places}.
\newblock \bibinfo{journal}{\emph{Computational and Mathematical Organization Theory}} \bibinfo{volume}{29}, \bibinfo{number}{1} (\bibinfo{year}{2023}), \bibinfo{pages}{20--51}.
\newblock


\end{thebibliography}

\end{document}